\documentclass[runningheads]{llncs}

\usepackage{styles/eccv}

\usepackage{styles/eccvabbrv}

\usepackage{graphicx}
\usepackage{booktabs}

\usepackage[accsupp]{axessibility}  

\usepackage{hyperref}

\usepackage{orcidlink}

\usepackage{booktabs}
\usepackage{colortbl}
\usepackage{xcolor}
\usepackage{array}
\usepackage{multirow}

\definecolor{avgcol}{RGB}{232,244,232}
\definecolor{bestcol}{RGB}{46,125,50}
\newcommand{\std}[1]{{\scriptsize$\pm$#1}}
\newcommand{\best}[1]{\textcolor{bestcol}{\textbf{#1}}}

\usepackage{pgfplots}
\makeatletter
\providecommand{\sf@counterlist}{}
\makeatother
\usepackage{subfig}

\usepackage{svg}
\svgpath{res/images/}

\usepackage[capitalize,noabbrev]{cleveref}

\usepackage{subfiles}

\usepackage{components/symbols}

\pgfplotsset{compat=1.18}

\begin{document}

\title{Multi-scale Mixture of World Models for Embodied Agents in Evolving Environments} 

\titlerunning{Multi-scale Mixture of World Models}

\author{Jinwoo Jang\inst{1}\orcidlink{0009-0003-8943-3565} \and
Daniel J. Rho\inst{1}\orcidlink{0009-0008-1448-0726} \and
Sihyung Yoon\inst{1}\orcidlink{0009-0005-3887-6299} \and
Hyunsuk Cho\inst{1}\orcidlink{0009-0004-1912-2654} \and
Honguk Woo\inst{1,2}\orcidlink{0000-0001-6948-3440}}

\authorrunning{Jang et al.}

\institute{Sungkyunkwan University
\email{\{jinustar,danielrho,godboy3752,hscho5133,hwoo\}@skku.edu} \and
Corresponding author}

\maketitle

\begin{abstract}

    Embodied agents operating in the real world require multi-scale reasoning and knowledge adaptation as conditions change.
    We identify two challenges in applying Mixture of Experts (MoE) to this setting: routing lacks an explicit notion of scale, preventing targeted updates at specific scales, and a uniform update policy cannot accommodate the different rates at which knowledge at each scale becomes outdated.
    We present \oursol, a framework that addresses both challenges through scale-aware world model mixture and evolution.
    A two-stage routing mechanism first maps experiential distance, a measure of situational novelty inspired by Construal Level Theory, to a weight over continuous scale space via a meta-router, then selects world models within the identified scale.
    For adaptation, scale-dependent forgetting rates allow low-scale knowledge to refresh rapidly while high-scale abstractions persist, and gated inter-scale transfer maintains coherence across the hierarchy.
    Experiments on EmbodiedBench and HAZARD show that \oursol improves over state-of-the-art baselines.
    
\end{abstract}

\section{Introduction}
    
    Embodied agents powered by vision-language models (VLMs) have achieved significant progress in complex instruction following~\cite{SayCanPay, FLARE, Huang2022InnerME}, yet real-world deployment demands multi-scale reasoning, from low-level physical dynamics to high-level abstract inference, while continuously adapting to non-stationary environments.
    Recent work has explored hierarchical world models that decompose predictions across scales~\cite{Cosmos-Reason1, THICK, STAR}, but these approaches do not address how world models should adapt when environmental conditions change over time.
    
\begin{figure*}[t]
    \centering
    \includegraphics[width=0.99\textwidth]{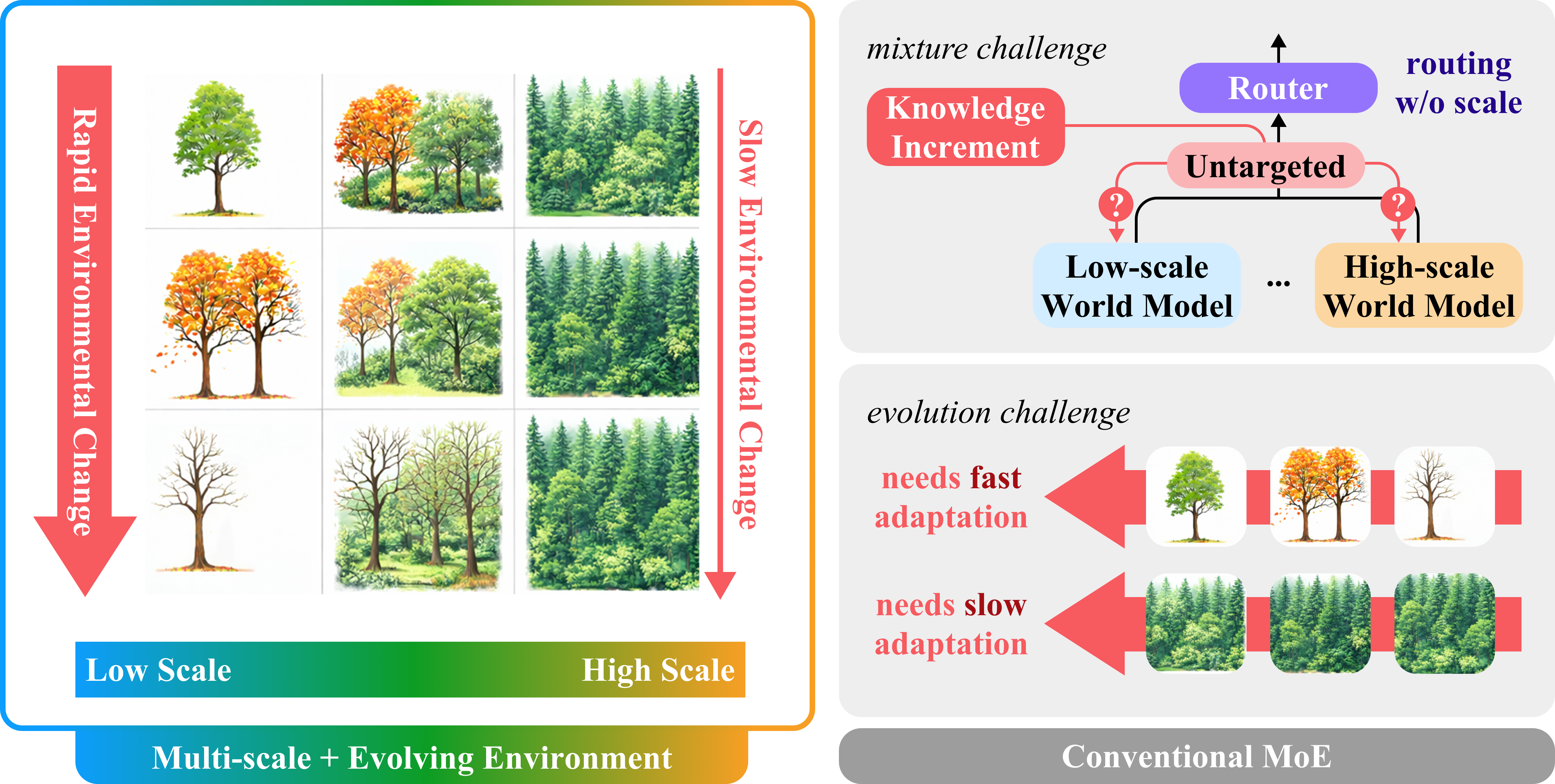}
    \caption{Explanation of mixture challenge and evolution challenge on conventional MoE}
    \vspace{-0.08in}
    \label{fig:introduction}
\end{figure*}

    Mixture of Experts (MoE)~\cite{MoE} offers a natural substrate for these demands: its modular expert selection can accommodate qualitatively different knowledge types, while its selective activation enables targeted updates without disrupting other components, effectively mitigating catastrophic forgetting~\cite{MoE-CL-Theory, MoE-RL}.
    However, conventional MoE is not scale-aware, introducing two limitations (Figure~\ref{fig:introduction}).
    First, standard routing operates without an explicit notion of scale, so world model selection is not tied to any identifiable scale, precluding test-time updates that target only the relevant scale (\textit{mixture challenge}).
    Second, a single uniform update policy cannot respect the fact that low-level knowledge about local dynamics changes frequently while high-level abstract rules remain relatively stable, preventing each scale from evolving at its own appropriate rate (\textit{evolution challenge}).

    To address these limitations, we propose \oursol, a multi-scale mixture of world models framework that enables embodied agents to dynamically mix and evolve world models at different scales.
    For the \textit{mixture challenge}, we introduce two-stage scale-based routing that explicitly separates scale determination from world model selection, yielding transparent routing in which the identified scale directly determines how knowledge increments are distributed across groups.
    For the \textit{evolution challenge}, we propose intra- and inter-scale knowledge adaptation mechanisms that allow each scale to evolve at its own characteristic rate while maintaining coherence across the hierarchy.
    To ground scale selection in a principled criterion, we draw on Construal Level Theory (CLT)~\cite{CLT}, which posits that psychological distance governs the level of abstraction in human reasoning, and operationalize this principle through experiential distance, a measure of how novel the current situation is relative to the agent's accumulated experience.
    
    To evaluate \oursol, we conduct experiments on EmbodiedBench~\cite{EmbodiedBench} and HAZARD~\cite{HAZARD}, covering both multi-scale reasoning and dynamic adaptation scenarios.
    EmbodiedBench evaluates agents across diverse capability dimensions including complex reasoning and spatial understanding, while HAZARD tests adaptation under evolving disaster conditions such as fire, flood, and wind.
    Our framework achieves 6.05\%p improvement over SayCanPay~\cite{SayCanPay} on EmbodiedBench (Habitat) and 1.49\%p improvement over FLARE~\cite{FLARE} on HAZARD (Fire), demonstrating effective multi-scale world model mixture and evolution. 
    
    Our contributions are as follows:
    \begin{itemize}
        \item We propose \oursol, a multi-scale mixture of world models framework for embodied agents that grounds world model selection in experiential distance, allowing agents to dynamically mix and evolve scale-specific world models at test time.
        \item We introduce two-stage scale-based routing that decomposes routing into scale determination via a continuous meta-router and world model selection via per-scale base routers, enabling scale-aware mixture and targeted test-time updates at the appropriate scale.
        \item We devise intra- and inter-scale knowledge adaptation mechanisms that respect the distinct temporal characteristics at each scale, enabling rapid adjustment of transient low-level details while preserving stable high-level abstractions through gated cross-scale transfer.
        \item We validate \oursol on EmbodiedBench and HAZARD benchmarks, demonstrating that scale-aware routing and knowledge adaptation yield 6.05\%p and 1.49\%p improvements over state-of-the-art baselines on multi-scale reasoning and dynamic adaptation scenarios, respectively. 
    \end{itemize}

\section{Related Works}

    \subsubsection{LM- and VLM-based embodied instruction following.}

        VLM-based embodied instruction following studies how an agent grounds natural-language instructions and vision observations in a physical environment through sequential action execution. Prior work has explored different strategies, including affordance-grounded approaches that combine LM reasoning with environmental feedback to evaluate feasible actions~\cite{Ahn2022DoAI, Huang2022InnerME}, and end-to-end vision-language models that directly map multimodal observations to action sequences~\cite{Driess2023PaLMEAE, zitkovich2023rt}. However, these approaches are often built around a fixed set of environmental assumptions or a monolithic model component, which hinders targeted adaptation to new environments and forces full system retraining even when only specific components need updating. Our \oursol framework addresses this by adopting a Mixture-of-Experts (MoE) design that decomposes world knowledge into modular components and selects among them during inference.

    \subsubsection{Mixture-of-Experts.}
    
        Mixture-of-Experts architectures have shown strong potential for domain adaptation, routing each input through a selected subset of specialized experts, allowing targeted updates to only the relevant experts during adaptation and thus mitigating catastrophic forgetting~\cite{Zhong2022MetaDMoEAT}. This routing paradigm has been extended to diverse settings, including vision-language adaptation~\cite{Shen2023ScalingVM} and multi-task learning~\cite{Ma2018ModelingTR}. However, conventional MoE routing is typically learned over latent representations and is often associated with specific tasks or domains, rather than an explicit notion of scale. As a result, it can be nontrivial to consistently invoke or adapt a desired knowledge granularity at test time without affecting other experts. Our approach addresses this by introducing experiential distance as an explicit, continuous scale axis that grounds routing decisions in a measurable quantity. This enables both interpretable world model selection and targeted test-time training that updates only the relevant scales.

    \subsubsection{Construal Level Theory (CLT).}

        Construal Level Theory (CLT)~\cite{CLT} posits that psychological distance determines the level of abstraction at which people mentally represent objects and events. Psychologically close entities elicit concrete, detail-rich representations (low-level construal), whereas distant entities elicit abstract, schematic representations (high-level construal). This distance-abstraction pattern has been documented across temporal, spatial, social, and hypothetical dimensions of psychological distance~\cite{ConstrualLA}.
        While CLT offers a well-established account of how distance modulates abstraction in human cognition, its potential as a measurable scale axis for routing mechanisms in MoE-based systems remains underexplored.
        Drawing on this correspondence, we define experiential distance as a continuous scalar that captures the gap between the agent’s accumulated experience and a given observation, and use it to govern hierarchical expert selection.
    
\section{\oursol} \label{sec:approach}
    
    \subsection{Overall Framework} \label{sec:overall}
    
\begin{figure*}[t]
    \centering
    \includegraphics[width=0.99\textwidth]{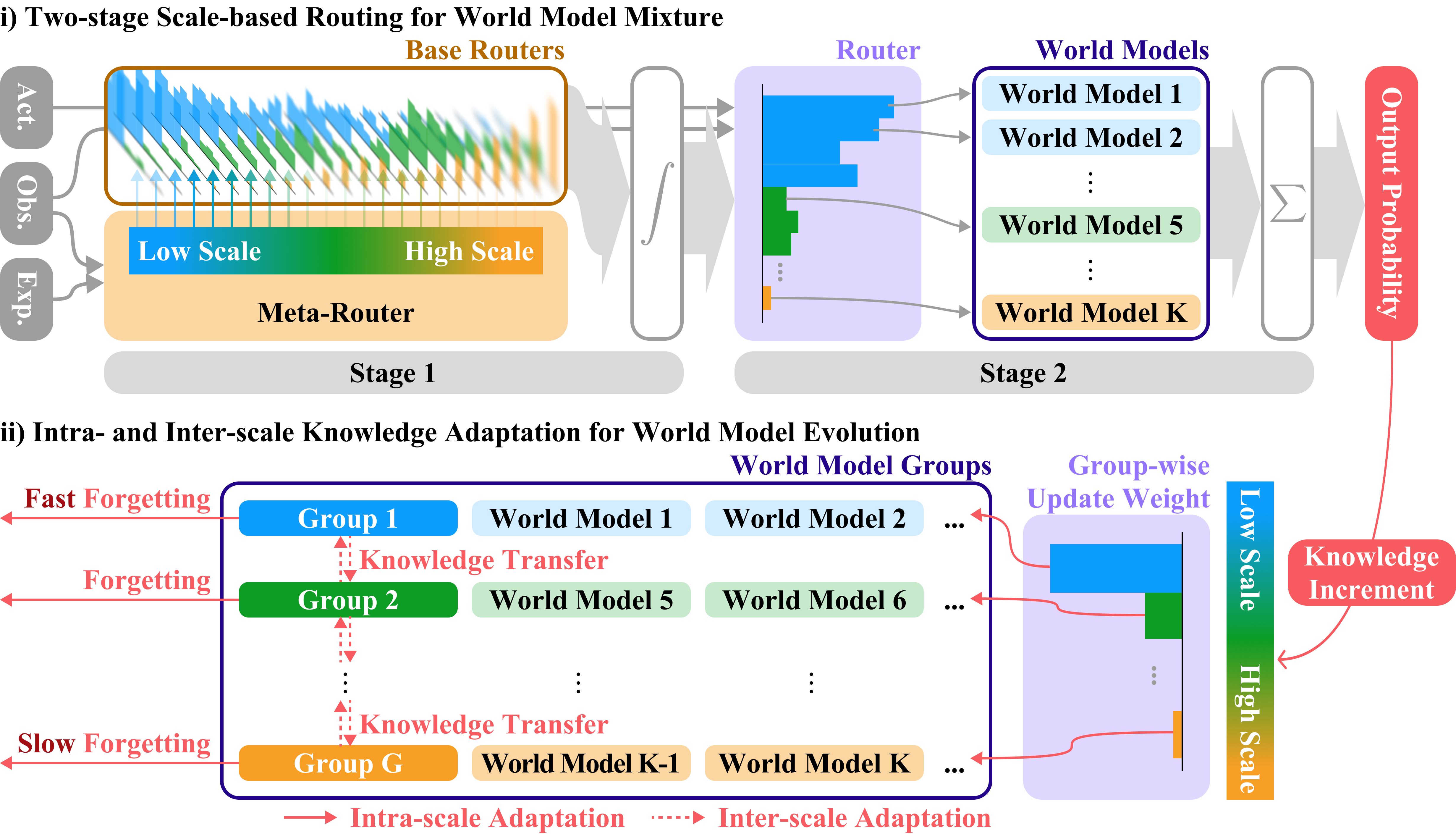}
    \caption{Overall framework of \oursol}
    \vspace{-0.08in}
    \label{fig:overall_framework}
\end{figure*}

        We present \oursol, a framework for embodied agents that integrates qualitatively different types of knowledge through scale-aware world model selection driven by experiential distance, a scalar measure of how novel the current situation is relative to the agent's accumulated experience.
        As illustrated in Figure~\ref{fig:overall_framework}, the framework addresses the two challenges identified in Section~1 through its core components.
        
        \textbf{Two-stage scale-based routing} (Section~\ref{sec:routing}) addresses the \textit{mixture challenge}.
        In the first stage, a meta-router maps experiential distance to a weight function over a continuous scale space.
        In the second stage, these scale weights are integrated with per-scale base router scores to produce the final router that selects world models.
        This decoupled design makes the scale of each routing decision explicit, enabling targeted adaptation at the appropriate scale.
        
        \textbf{Intra- and inter-scale knowledge adaptation} (Section~\ref{sec:adaptation}) addresses the \textit{evolution challenge}.
        Intra-scale adaptation distributes knowledge increments to the appropriate scale via the meta-router and applies scale-dependent forgetting rates so that each level evolves at its own pace.
        Inter-scale adaptation then extends this with gated cross-scale knowledge transfer to maintain coherence across the hierarchy.

    \subsection{Two-stage Scale-based Routing for World Model Mixture} \label{sec:routing}
    
        \subsubsection{World model groups and mixture of world models.}
            
            Our framework organizes world models into $\ngroup$ groups, ranging from low-scale processing to high-scale reasoning.
            Each group is associated with a distinct architecture type; for example, an RBF Network can serve as a low-scale world model, while a Cognitive Map can act as a high-scale world model.
            Here, the world model is a module $\wm : (\obs_t, \act_t) \mapsto \obs_{t+1}$, trained via teacher-forcing on a dataset $\mathcal D = \{(\text{task}_i, \traj_i)\}_{i=1}^D$ of $D$ task-trajectory pairs, where $\traj_i = \{(\obs_t, \act_t, \obs_{t+1})\}_{t=1}^T$.
            The framework comprises $\nwm$ world models $\{\wm_1, \ldots, \wm_\nwm\}$ distributed across $\ngroup$ groups and selects a top-$k$ subset for each input to form the mixture of world models $M$:
            \begin{equation}
                M = \sum_{i=1}^\nwm\!\wrouter{i} \wm_i; \quad
                \wrouter{i} = \softmax(\topk{k}(\router(\obs_t, \act_t)))_i,
            \end{equation}
            where $\obs_t \in \obsspace$ and $\act_t \in \actspace$ denote the current observation and action, $\obsspace$ and $\actspace$ are the observation space and the action space, respectively, and $\router : \obsspace \times \actspace \rightarrow \mathbb{R}^\nwm$ is the router that determines routing weights for each world model.
        
        \subsubsection{Continuous meta-router and two-stage routing.}
            
            A single flat router cannot identify which scale is currently active, making it impossible to selectively adapt world model groups at the relevant scale during test time.
            To address this, we decompose routing into two stages: a \textit{continuous meta-router} $\mr$ that first determines scale weights based on experiential distance, and a final \textit{router} $\router$ that integrates these weights with per-scale \textit{base router} scores $r(\s)$, $\s \in \sspace$, to select world models.
            
            The meta-router produces a weight function over the scale space, and the router integrates over all scales:
            \begin{equation}
                \router = \int_\sspace \wmr(\s) r(\s) \, d\s; \quad
                \wmr(\s) = \softmax_\sspace(\mr(\obs_t, \ex_{<t}, \s)),
            \end{equation}
            where $\softmax_\sspace$ denotes the continuous softmax over $\sspace$,
            \begin{equation}
                \softmax_\sspace(f(\s)) = \frac{\exp(f(\s))}{\int_\sspace \exp(f(\s')) \, d\s'},
            \end{equation}
            and $\mr : \obsspace \times \obsspace^* \times \sspace \rightarrow \mathbb R$ is the continuous meta-router.
            $\sspace$ is a compact subset of a normed vector space with $R_\sspace = \sup_{\s \in \sspace} \lVert \s \rVert$, and $\ex_{<t} = (\obs_1, \ldots, \obs_{t-1}) \in \obsspace^*$ denotes the sequence of observations accumulated within the episode, encoding the agent's past experience.
            In practice, we approximate the integrals via Monte Carlo sampling over $\sspace$.
            
            For each scale $\s \in \sspace$, the base router $r(\s) : \obsspace \times \actspace \rightarrow \mathbb{R}^\nwm$ outputs unnormalized world model routing scores, with lower scales favoring low-scale world models and higher scales favoring high-scale world models.
            
        \subsubsection{Experiential distance and scale-aware loss.} \label{sec:exdist}
            
            The meta-router conditions on experiential distance $\exdist$, which quantifies the deviation of the current situation from the agent's accumulated experience.
            Let $\boldsymbol\phi : \obsspace \rightarrow \mathbb{R}^d$ be a frozen pretrained encoder.
            Recall that experience $\ex_{<t} = (\obs_1, \ldots, \obs_{t-1})$ is the observation sequence accumulated within an episode.
            We model its embedding distribution as a multivariate Gaussian.
            To ensure that the distribution reflects the agent's most recent experience, we assign exponentially decaying weights $w_i = \exp(-\beta(t - i))$, where $\beta > 0$ controls the decay rate.
            The weighted mean and covariance in the embedding space are:
            \begin{equation}
                \boldsymbol{\mu}_\ex = \frac{\sum_{i=1}^{t-1} w_i \boldsymbol\phi(\obs_i)}{\sum_{i=1}^{t-1} w_i}, \quad
                \boldsymbol{\Sigma}_\ex = \frac{\sum_{i=1}^{t-1} w_i (\boldsymbol\phi(\obs_i) - \boldsymbol{\mu}_\ex)(\boldsymbol\phi(\obs_i) - \boldsymbol{\mu}_\ex)^\top}{\sum_{i=1}^{t-1} w_i}.
            \end{equation}
            The experiential distance is then defined as the Mahalanobis distance between the current observation embedding and this distribution:
            \begin{equation}
                \exdist(\obs_t, \ex_{<t}) = \sqrt{(\boldsymbol\phi(\obs_t) - \boldsymbol{\mu}_\ex)^\top \boldsymbol{\Sigma}_\ex^{-1} (\boldsymbol\phi(\obs_t) - \boldsymbol{\mu}_\ex)}.
            \end{equation}
            When $t < T_{\min}$, where $T_{\min}$ is the minimum number of steps required to estimate $\boldsymbol{\Sigma}_\ex$, we set $\exdist = +\infty$, reflecting that all situations are novel without prior experience.
            
            To align the meta-router with experiential distance, we first normalize $\exdist$ to a bounded range via $\bar{\exdist} = 1 - \exp(-\tau \exdist)$, mapping $\exdist \in [0, \infty)$ to $[0, 1)$, where $\tau > 0$ controls the transition sharpness.
            The \textit{scale-aware loss} is then:
            \begin{equation}
                \mathcal{L}_\sspace = \underbrace{(R_\sspace \bar{\exdist} - \lVert \boldsymbol{\mu}_\sspace \rVert)^2}_{\mathcal{L}_\sspace^{(1)}} + \underbrace{\lambda_1 \mathrm{tr}(\boldsymbol{\Sigma}_\sspace)}_{\mathcal{L}_\sspace^{(2)}} + \underbrace{\lambda_2 Z^2}_{\mathcal{L}_\sspace^{(3)}},
                \label{eq:scale_loss}
            \end{equation}
            where $\boldsymbol{\mu}_\sspace$ and $\boldsymbol{\Sigma}_\sspace$ are the weighted mean and covariance of $\s$ under $\wmr$, and $\lambda_1, \lambda_2 > 0$ are balancing coefficients.
            $Z = \int_\sspace \mr(\obs, \ex, \s) d\s$ is the integral of the meta-router output.
            The first term ($\mathcal L_\sspace^{(1)}$) encourages the expected scale magnitude to match the experiential distance, routing familiar situations to lower scales and novel situations to higher scales.
            The second term ($\mathcal L_\sspace^{(2)}$) encourages the scale distribution to concentrate rather than spread diffusely across the scale space.
            The third term ($\mathcal L_\sspace^{(3)}$) prevents the meta-router outputs from diverging.
            
            The overall training objective combines world model prediction loss with scale alignment:
            \begin{equation}
                \mathcal{L} = \mathcal{L}_{\mathrm{TF}} + \lambda_\sspace \mathcal{L}_\sspace,
            \end{equation}
            where $\mathcal{L}_{\mathrm{TF}}$ is the teacher-forcing loss that trains the model to predict next action or observation from the trajectory~\cite{traj-transformer,unified-world-models} and $\lambda_\sspace > 0$ controls the strength of scale alignment.
            
            The design of experiential distance and the scale-aware loss draws on Construal Level Theory (CLT) in cognitive science~\cite{CLT}, which posits that psychological distance governs the level of abstraction in human reasoning.
            CLT further posits that this abstraction arises from multiple dimensions, including spatial, temporal, social, and hypothetical distance; our multi-dimensional scale space reflects this structure.
            Applying this principle, the scale-aware loss routes low-$\exdist$ situations to low-scale world models that operate on concrete, well-established patterns, and high-$\exdist$ situations to high-scale world models capable of abstract generalization.
        
    \subsection{Intra- and Inter-scale Knowledge Adaptation for World Model Evolution} \label{sec:adaptation}
    
        To enable selective adaptation at each scale, we introduce the intra- and inter-scale knowledge adaptation mechanism that updates world models at test time.
        We define knowledge states $\{\knowledge^{(1)}, \ldots, \knowledge^{(\ngroup)}\}$, where $\knowledge^{(\group)} \in \knowledgespace^*$ represents the accumulated knowledge for group $\group$ (e.g. neural memory), and $\knowledgespace$ is a knowledge space.
        Each group-$\group$ world model exclusively references its corresponding knowledge state $\knowledge^{(\group)}$, maintaining clear separation of scale-specific knowledge.
        
        During inference, the prediction error yields a knowledge increment $\Delta\knowledge_{t+1}$ at each step $t \rightarrow t+1$ (in the spirit of the surprise metric in \cite{behrouz2026titans}). This increment is distributed across scales via the meta-router and integrated with scale-dependent forgetting (\textit{intra-scale adaptation}), and then propagated through gated cross-scale transfer (\textit{inter-scale adaptation}).
        
        \subsubsection{Intra-scale adaptation.}
        
            We distribute each knowledge increment to the appropriate scale via the meta-router and apply scale-dependent forgetting, so that each level adapts at a rate matching the volatility of its content.
            
            Given a knowledge increment $\Delta\knowledge_{t+1}$, the agent must first determine which scale requires updating.
            Our two-stage routing enables this: given the next observation $\obs_{t+1}$, the meta-router identifies the current scale as
            \begin{equation}
                \s^* = \argmax{\s \in \sspace}\ \mr(\obs_{t+1}, \ex_{<t+1}, \s).
            \end{equation}
            The knowledge increment is then distributed across groups according to the group-wise update weight $\eta^{(\group)}$:
            \begin{equation}
                \Delta\knowledge_{t+1}^{(\group)} = \eta^{(\group)} \Delta\knowledge_{t+1}; \quad
                \eta^{(\group)} = \sum_{i=1}^\nwm \mathbf{1}[\wm_i \in \text{group } \group] \cdot \softmax(r(\s^*)(\obs_t, \act_t))_i,
            \end{equation}
            ensuring that only the relevant scale receives substantial updates.
            
            Each group then integrates its increment with scale-dependent forgetting:
            \begin{equation} \label{eq:intra_update}
                \knowledge_{t+1}^{(\group)} = (1 - \alpha^{(\group)}) \knowledge_{t}^{(\group)} + \Delta\knowledge_{t+1}^{(\group)},
            \end{equation}
            where $\alpha^{(\group)} = \alpha_{\max} \exp(-\gamma (\group-1))$, with $\alpha_{\max}$ controlling the maximum forgetting rate and $\gamma > 0$ determining the decay across groups.
            Low-scale groups exhibit rapid forgetting ($\alpha^{(\group)}$ large) while high-scale groups retain information over extended periods ($\alpha^{(\group)}$ small).
        
        \subsubsection{Inter-scale adaptation.}
        
            We extend the isolated per-group update (Equation~\ref{eq:intra_update}) with gated bidirectional knowledge transfer across neighboring scales, so that knowledge gained at one level can inform adjacent levels.
            Gating parameters $\mathbf W_+^{(\group)}, \mathbf W_-^{(\group)}$ modulate information flow between neighboring groups:
            \begin{equation}
                g_{+}^{(\group)} = \sigma(\boldsymbol\phi(\obs_{t+1}) \mathbf W_+^{(\group)}),\quad g_{-}^{(\group)} = \sigma(\boldsymbol\phi(\obs_{t+1})\mathbf  W_-^{(\group)}),
            \end{equation}
            where $\mathbf W_+^{(\group)}$ and $\mathbf W_-^{(\group)}$ are learned during training and held fixed at test time, while only the knowledge states $\knowledge^{(\group)}$ are updated.
            With cross-scale transfer, the update rule in Equation~\ref{eq:intra_update} extends to:
            \begin{equation}
                \knowledge_{t+1}^{(\group)} = (1 - \alpha^{(\group)}) \knowledge_{t}^{(\group)} + \Delta\knowledge_{t+1}^{(\group)} + \Delta\knowledge_{\mathrm{in},t+1}^{(\group)} - \Delta\knowledge_{\mathrm{out},t+1}^{(\group)},
            \end{equation}
            where
            \begin{align}
                \Delta\knowledge_{\mathrm{in},t+1}^{(\group)} &= g_{+}^{(\group)} \odot \knowledge_{t}^{(\group-1)} + g_{-}^{(\group)} \odot \knowledge_{t}^{(\group+1)}, \\
                \Delta\knowledge_{\mathrm{out},t+1}^{(\group)} &= g_{+}^{(\group+1)} \odot \knowledge_{t}^{(\group)} + g_{-}^{(\group-1)} \odot \knowledge_{t}^{(\group)}.
            \end{align}
            Terms involving $\knowledge^{(0)}$, $\knowledge^{(\ngroup+1)}$, $g_{+}^{(\ngroup+1)}$, or $g_{-}^{(0)}$ are set to zero at boundaries.
            The incoming term $\Delta\knowledge_{\mathrm{in},t+1}^{(\group)}$ aggregates knowledge from neighboring groups, while the outgoing term $\Delta\knowledge_{\mathrm{out},t+1}^{(\group)}$ accounts for knowledge transferred away, ensuring consistency across the hierarchy.
    
\section{Experiments} \label{sec:experiments}

    We evaluate \oursol on two complementary benchmarks: EmbodiedBench~\cite{EmbodiedBench} for multi-scale reasoning capabilities and HAZARD~\cite{HAZARD} for adaptation in dynamic environments.

    \subsection{Experimental Setup} \label{sec:setup}
    
        \subsubsection{Benchmarks.}
            
            EmbodiedBench evaluates embodied agents across six capability dimensions in two simulation environments.
            EB-Habitat tests household task completion requiring object manipulation and navigation, while EB-Navigation focuses on indoor navigation with realistic visual observations.
            HAZARD evaluates agents in dynamic disaster scenarios (fire, flood) where environmental conditions evolve over time, requiring continuous adaptation of both low-level reactive behaviors and high-level rescue strategies.
        
        \subsubsection{Baselines.}
            
            We compare against representative methods spanning different para-digms:
            \textbf{LLM-Planner}~\cite{LLM-Planner} generates plans through few-shot prompting without explicit world modeling.
            \textbf{SayCanPay}~\cite{SayCanPay} combines language model planning with learned affordance functions.
            \textbf{FLARE}~\cite{FLARE} employs retrieval-augmented generation for grounded action prediction.
            For ablation, we include a \textbf{Conventional MoE} variant that uses standard top-$k$ routing without scale-aware mechanisms.
        
        \subsubsection{Implementation details.}
            
            We use Qwen3-VL-4B-Instruct~\cite{Qwen3-VL} as the vision-language backbone.
            Our framework consists of $G=3$ world model groups with $K=15$ world models distributed across scales, and operates in a 3-dimensional scale space.
            We select top-4 world models per input.
            For knowledge adaptation, we set $\alpha_{\max}=0.3$ and $\gamma=1.0$.
            Hyperparameters are selected via grid search.
        
    \subsection{Main Results} \label{sec:main_results}
    
        \subsubsection{Multi-scale reasoning on EmbodiedBench.}
        
            \begin{table}[t]
    \centering
    \caption{\textbf{Evaluation result on EB-Habitat and EB-Navigation.} We report success rate (\%) for each capability subset within EB-Habitat and EB-Navigation environments.}
    \label{tab:embodiedbench_1}
    \vspace{2mm}
    \footnotesize
    \setlength{\tabcolsep}{3pt}
    \centering
    \resizebox{\textwidth}{!}{
        \begin{tabular}{@{}lcccccc>{\columncolor{avgcol}}cccccc>{\columncolor{avgcol}}c@{}}
        \toprule
        \multirow{2}{*}[-0.5ex]{\textbf{Methods}} & \multicolumn{7}{c}{\textbf{EB-Habitat}} & \multicolumn{6}{c}{\textbf{EB-Navigation}} \\
        \cmidrule(lr){2-8} \cmidrule(lr){9-14}
        & Base & Cmn & Cpx & Vis & Spt & Lng & Avg ($\uparrow$) & Base & Cmn & Cpx & Vis & Lng & Avg ($\uparrow$) \\
        \midrule
        LLM-Planner & 53.33 & 2.67 & 18.67 & 22.67 & 30.67 & 10.67 & 23.11\std{1.45} & 36.80 & 30.40 & 27.20 & 28.80 & 1.60 & 24.96\std{0.67} \\
        SayCanPay & 72.00 & 11.67 & 30.67 & 37.33 & \best{37.33} & 17.33 & 34.39\std{2.46} & 54.40 & 47.20 & \best{57.60} & 48.00 & 28.80 & 47.20\std{0.98} \\
        FLARE & 72.00 & 4.00 & 18.67 & 29.33 & 28.00 & 10.67 & 27.11\std{3.48} & 45.60 & 35.20 & 44.00 & 32.00 & 14.40 & 34.24\std{3.64} \\
        Conventional MoE & 60.00 & 10.67 & 8.00 & 32.00 & 30.67 & 12.00 & 25.56\std{1.36} & 36.00 & 28.80 & 39.20 & 32.80 & 9.60 & 29.28\std{2.50} \\
        \midrule
        \textbf{\oursol} & \best{73.33} & \best{26.67} & \best{37.33} & \best{42.67} & 36.00 & \best{26.67} & \best{40.44}\std{1.27} & \best{63.20} & \best{63.20} & \best{57.60} & \best{62.40} & \best{43.20} & \best{57.92}\std{1.75} \\
        \bottomrule
        \end{tabular}
    }
\end{table}
            
            Table~\ref{tab:embodiedbench_1} presents results across six capability dimensions.
            \oursol achieves consistent improvements over baselines in both EB-Habitat and EB-Navigation environments.
            Notably, our method shows substantial gains on tasks requiring complex reasoning (Cpx) and visual understanding (Vis), which demand effective integration of abstract planning with concrete physical inference.
            The improvement over Conventional MoE demonstrates that scale-aware routing is essential for leveraging the multi-scale architecture effectively.
            
        \subsubsection{Adaptation in dynamic environments on HAZARD.}
        
            \begin{table*}[!tb]
    \centering
    \caption{\textbf{Evaluation result on HAZARD benchmark.} We report value of rescued objects (Val), number of steps (Step), and damage ratio (Dmg) across three dynamic disaster scenarios. We use 3 random seeds.}
    \label{tab:hazard}
    \vspace{2mm}
    \footnotesize
    \setlength{\tabcolsep}{4pt}
    \centering
    \resizebox{0.99\textwidth}{!}{
        \begin{tabular}{@{}lccc|ccc@{}}
        \toprule
        \multirow{2}{*}[-0.5ex]{\textbf{Methods}} & \multicolumn{3}{c|}{\textbf{Fire}} & \multicolumn{3}{c}{\textbf{Flood}} \\
        \cmidrule(lr){2-4} \cmidrule(lr){5-7}
        & Val ($\uparrow$) & Step ($\downarrow$) & Dmg ($\downarrow$) & Val ($\uparrow$) & Step ($\downarrow$) & Dmg ($\downarrow$) \\
        \midrule
        LLM-Planner & 38.48\std{7.91} & 1285.42\std{272.72} & 2.53\std{0.13} & 49.78\std{0.07} & 1145.20\std{7.46} & 6.60\std{0.35} \\
        SayCanPay & 41.07\std{1.43} & 1467.72\std{12.50} & 2.55\std{0.11} & \best{49.83}\std{0.00} & 1148.17\std{2.56} & 6.80\std{0.50} \\
        FLARE & 42.22\std{0.75} & 1480.48\std{7.14} & 2.88\std{0.32} & 49.00\std{1.18} & 1133.90\std{23.10} & 6.71\std{0.46} \\
        Conventional MoE & 41.53\std{0.83} & 1500.77\std{7.73} & 2.89\std{0.12} & \best{49.83}\std{0.00} & 1123.30\std{18.03} & 7.49\std{0.68} \\
        \midrule
        \textbf{\oursol} & \best{43.71}\std{0.82} & \best{1466.2}\std{17.90} & \best{1.95}\std{0.45} & \best{49.83}\std{0.00} & \best{1130.7}\std{12.70} & \best{4.65}\std{0.60} \\
        \bottomrule
        \end{tabular}
    }
\end{table*}
            
            Table~\ref{tab:hazard} shows results on disaster scenarios with evolving environmental conditions.
            In the fire scenario, \oursol achieves the highest rescue value among all methods.
            In the flood scenario, rescue values are comparable across methods within error margins, but \oursol notably reduces the damage ratio ($4.65$ vs. $6.60$--$7.49$ for baselines), indicating more efficient rescue behavior that avoids unnecessary environmental exposure.
            These results suggest that multi-scale routing and knowledge adaptation enable the agent to balance quick low-level reactions with stable high-level strategies as conditions evolve.
        
        \subsubsection{Real-world robotic manipulation.}
        
            \begin{table}[t]
    \centering
    \caption{\textbf{Real-world robotic manipulation with Franka Research 3.} We report success rate (\%) across 8 manipulation tasks.
    }
    \label{tab:realworld}
    \vspace{2mm}
    \footnotesize
    \setlength{\tabcolsep}{6pt}
    \resizebox{0.99\textwidth}{!}{
        \begin{tabular}{@{}lcccccccc>{\columncolor{avgcol}}c@{}}
        \toprule
        \textbf{Method} & \textbf{Task 1} & \textbf{Task 2} & \textbf{Task 3} & \textbf{Task 4} & \textbf{Task 5} & \textbf{Task 6} & \textbf{Task 7} & \textbf{Task 8} & \textbf{Avg} ($\uparrow$) \\
        \midrule
        LLM-Planner & 66.7 & 16.7 & 33.3 & 7.6 & 50.0 & 0.0 & 0.0 & 0.0 & 21.8\std{2.0} \\
        SayCanPay & 66.7 & 66.7 & 83.3 & 33.3 & 83.3 & 66.7 & 17.5 & 50.0 & \best{58.4}\std{0.9} \\
        FLARE & 66.7 & 16.7 & 33.3 & 28.5 & 50.0 & 0.0 & 0.0 & 0.0 & 24.4\std{2.8} \\
        Conventional MoE & 100.0 & 100.0 & 0.0 & 33.3 & 73.3 & 70.0 & 30.0 & 23.3 & 53.7\std{2.7} \\
        \midrule
        \textbf{\oursol} & 66.7 & 50.0 & 83.3 & 50.0 & 76.7 & 66.7 & 30.0 & 26.7 & \best{56.2}\std{1.9} \\
        \bottomrule
        \end{tabular}
    }
\end{table}
            \begin{figure}[!tb]
    \centering
    \begin{subfigure}[t]{0.48\textwidth}
        \centering
        \centering
        \includegraphics[width=0.95\textwidth]{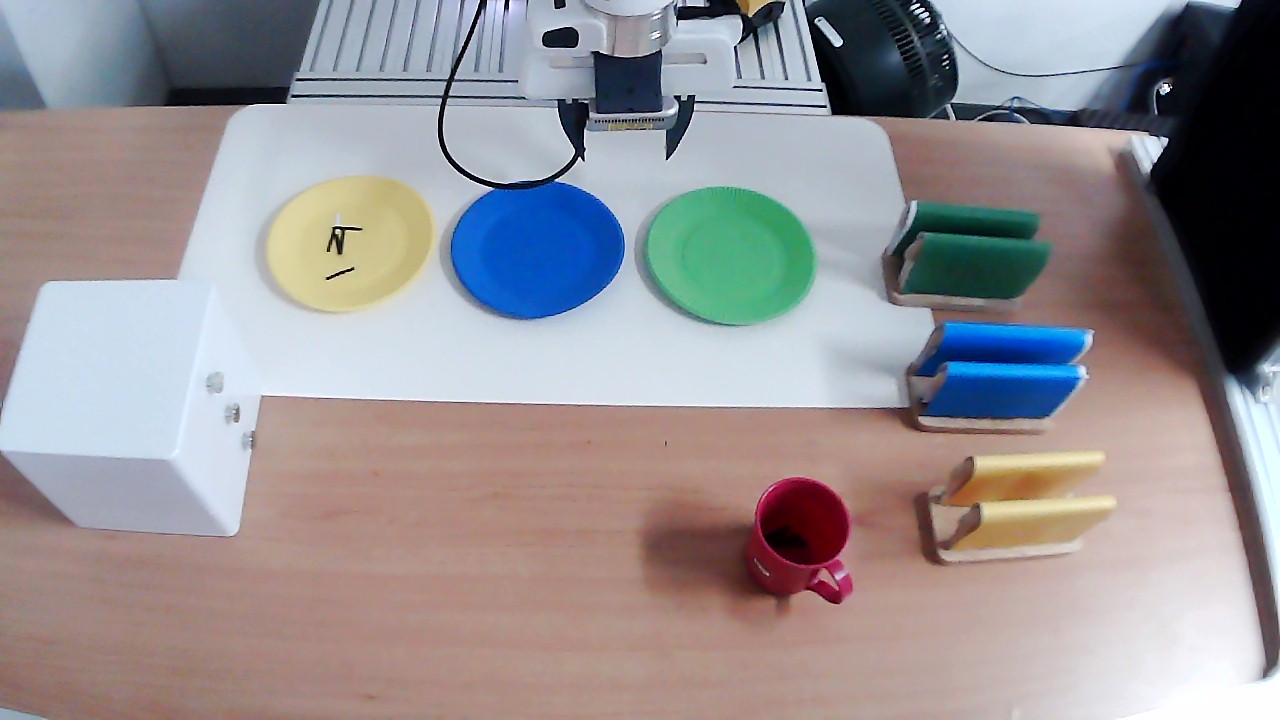}
        \label{fig:real_world_a}
    \end{subfigure}
    \hfill
    \begin{subfigure}[t]{0.48\textwidth}
        \centering
        \includegraphics[width=0.95\textwidth]{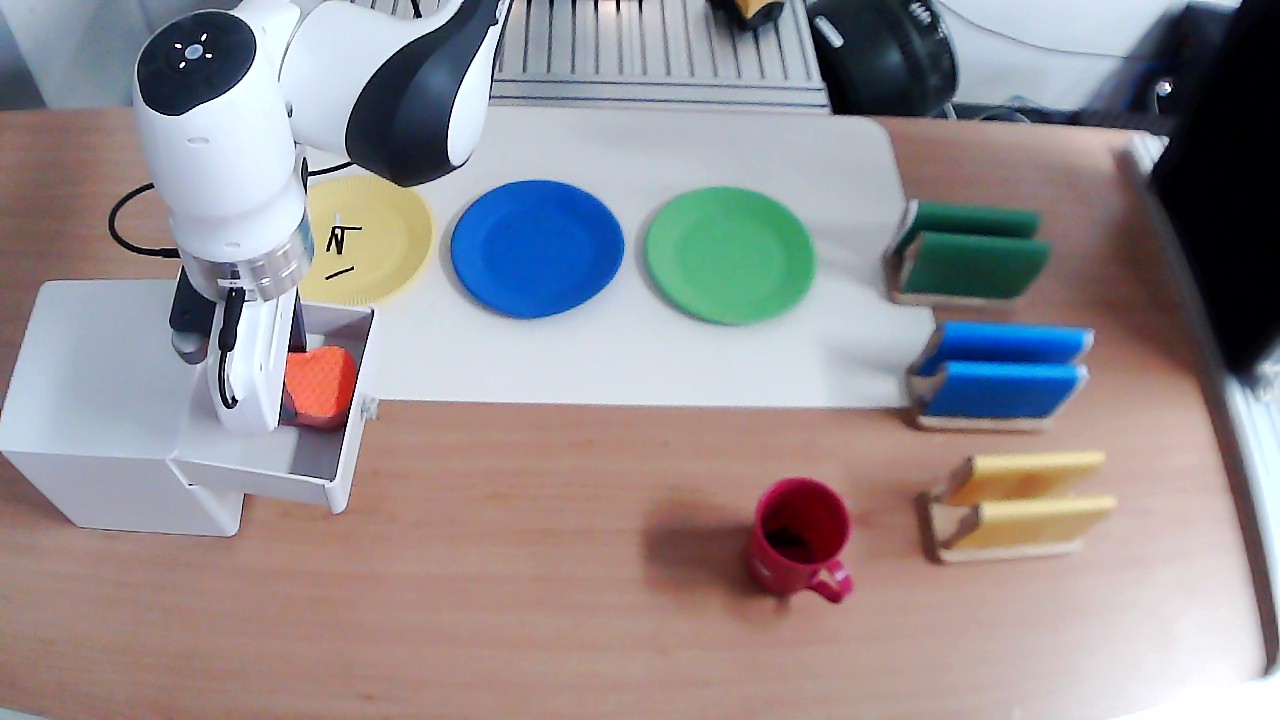}
        \label{fig:real_world_b}
    \end{subfigure}
    \caption{Real-world robotic manipulation examples}
    \label{fig:real_world_examples}
\end{figure}
        
            To validate the practical applicability of \oursol, we conduct real-world experiments using a Franka Research 3 robot arm (Figure~\ref{fig:real_world_examples}), evaluating across 8 manipulation tasks (Table~\ref{tab:realworld}).
            \oursol outperforms most baselines and performs comparably to the strongest one, SayCanPay, in average success rate.
            Moreover, \oursol attains the highest worst-case performance across tasks, indicating more consistent behavior under diverse real-world conditions.
        
    \subsection{Ablation Study} \label{sec:ablation}
        
        \begin{table}[t]
    \centering
    \caption{\textbf{Ablation study.} We report success rate (\%) on EmbodiedBench and rescue value on HAZARD.}
    \label{tab:ablation}
    \vspace{2mm}
    \footnotesize
    \setlength{\tabcolsep}{3pt}
    \resizebox{0.99\textwidth}{!}{
        \begin{tabular}{@{}lccc@{}}
        \toprule
        \multirow{2}{*}[-0.5ex]{\textbf{Method}} & \multicolumn{1}{c}{\textbf{EmbodiedBench (SR $\uparrow$)}} & \multicolumn{2}{c}{\textbf{HAZARD (Value $\uparrow$)}} \\
        \cmidrule(lr){2-2} \cmidrule(lr){3-4}
        & EB-Habitat & Fire & Flood \\
        \midrule
        \textbf{\oursol} (full) & \best{40.44} & 43.71 & \best{49.83}  \\
        \midrule
        \multicolumn{4}{@{}l}{\textit{Two-stage Scale-based Routing}} \\
        \quad w/o meta-router & 32.22 & 42.22 & \best{49.83} \\
        \quad w/o $\mathcal{L}_\sspace^{(1)}$ (alignment) & 37.78 & 39.85 & 45.38 \\
        \quad w/o $\mathcal{L}_\sspace^{(2)}$ (variance) & 35.56 & \best{44.89} & \best{49.83} \\
        \quad w/o $\mathcal{L}_\sspace^{(3)}$ (magnitude) & 38.89 & 44.52 & \best{49.83} \\
        \midrule
        \multicolumn{4}{@{}l}{\textit{Intra- and Inter-scale Knowledge Adaptation}} \\
        \quad w/o intra-scale adaptation & 38.89 & 39.84 & 45.38 \\
        \quad w/o inter-scale adaptation & 37.78 & 41.67 & 45.38 \\
        \bottomrule
        \end{tabular}
    }
\end{table}
        
        We conduct ablation studies to validate the contribution of each component.
        Table~\ref{tab:ablation} shows all ablations evaluated on both EmbodiedBench and HAZARD.
        
        \subsubsection{Ablations of two-stage scale-based routing.}

            Removing the meta-router (w/o meta-router) causes the largest overall degradation, with EB-Habitat dropping from $40.44$ to $32.22$, confirming that the two-stage hierarchy is essential for effective scale-aware routing.
            Without the meta-router, the current scale $\s^*$ cannot be identified, so the scale-group weight for knowledge adaptation falls back to $\eta^{(g)} = \sum_{i \in \text{group } g} w_{R,i}$, which simply aggregates routing weights without any explicit scale information.
            This validates the core motivation of two-stage routing, where identifying the current scale enables selective adaptation of world model groups at the relevant scale, a capability that scale-agnostic approaches cannot achieve.
        
            We further ablate each term of the scale-aware loss $\mathcal{L}_\sspace$ (Equation~\ref{eq:scale_loss}) to assess its individual contribution.
            Removing the alignment term (w/o $\mathcal{L}_\sspace^{(1)}$) leads to degradation across all benchmarks, confirming that explicit alignment between experiential distance and scale is necessary for the meta-router to learn scale-aware routing.
            The variance and magnitude regularizers (w/o $\mathcal{L}_\sspace^{(2)}$, $\mathcal{L}_\sspace^{(3)}$), in contrast, show environment-dependent effects.
            They improve stability on EB-Habitat but offer only marginal or mixed impact on HAZARD, suggesting that their inclusion should be tuned per deployment environment.
        
        \subsubsection{Ablations of intra- and inter-scale knowledge adaptation.}
        
            Disabling scale-dependent forgetting (w/o intra-scale) and applying uniform forgetting rates across all scales is harmful, as this variant either forgets stable high-scale knowledge too quickly or retains outdated low-scale details too long.
            Removing cross-scale knowledge transfer (w/o inter-scale) also results in consistent degradation.
            Without bidirectional information flow, knowledge remains siloed within each scale, preventing coordinated adaptation across the hierarchy.

    \subsection{Analysis} \label{sec:analysis}

        \subsubsection{World model group variants.}
            
            To verify that \oursol is not tied to a specific set of world model architectures, we evaluate five group configurations (Variants I--V) on EmbodiedBench, varying both the number of groups and architecture types. Detailed compositions are provided in Table~\ref{tab:group_compositions}.
            Low-scale groups employ architectures suited to modeling physical dynamics and local patterns, mid-scale groups capture relational and spatial structure, and high-scale groups handle abstract concept-level or rule-based reasoning.

            \begin{table}[!tb]
    \centering
    \caption{\textbf{World model group compositions.} Each variant assigns a specific architecture type to each group, ordered from low-scale (Group~1) to high-scale (Group~$\ngroup$).}
    \label{tab:group_compositions}
    \footnotesize
    \resizebox{\textwidth}{!}{
        \begin{tabular}{@{}llllll@{}}
            \toprule
            \textbf{Variant} & \textbf{$\ngroup$} & \textbf{Group 1} & \textbf{Group 2} & \textbf{Group 3} & \textbf{Group 4} \\
            \midrule
            I (default) & 3 & PINN              & Cognitive Map       & Relational Network    & --  \\
            II          & 3 & RBF Network       & Concept Bottleneck  & Schema Network        & --  \\
            III         & 4 & Sensory           & PINN                & Relational Network    & Concept Bottleneck \\
            IV          & 4 & Sensory           & RBF Network         & Cognitive Map         & Schema Network \\
            V           & 2 & PINN              & Schema Network      & --                    & --  \\
            \bottomrule
        \end{tabular}
    }
    \vspace{2mm}
\end{table}
            
            Table~\ref{tab:group_variants} shows that all five variants fall within a narrow range (38.89--42.22\% on EB-Habitat, 54.13--57.92\% on EB-Navigation), and even the weakest surpasses the strongest baseline (SayCanPay; 34.39\% and 47.20\%; Table~\ref{tab:embodiedbench_1}) by a clear margin.
            This confirms that the framework's effectiveness stems from the scale-aware routing mechanism rather than from a particular choice of world model architectures.
            
            \begin{table}[t]
    \centering
    \caption{\textbf{World model group variants on EmbodiedBench.} We report success rate (\%) for each capability subset. Variant~I is our default configuration. We use 3 random seeds.
    }
    \label{tab:group_variants}
    \footnotesize
    \setlength{\tabcolsep}{3pt}
    \centering
    \resizebox{\textwidth}{!}{
        \begin{tabular}{@{}lccccccc>{\columncolor{avgcol}}cccccc>{\columncolor{avgcol}}c@{}}
        \toprule
        \multirow{2}{*}[-0.5ex]{\textbf{Variant}} & \multirow{2}{*}[-0.5ex]{\textbf{$\ngroup$}} & \multicolumn{7}{c}{\textbf{EB-Habitat}} & \multicolumn{6}{c}{\textbf{EB-Navigation}} \\
        \cmidrule(lr){3-9} \cmidrule(lr){10-15}
        & & Base & Cmn & Cpx & Vis & Spt & Lng & Avg ($\uparrow$) & Base & Cmn & Cpx & Vis & Lng & Avg ($\uparrow$) \\
        \midrule
        I (default) & 3 & 73.33 & 26.67 & 37.33 & 42.67 & 36.00 & 26.67 & 40.44\std{1.27} & 63.20 & 63.20 & 57.60 & 62.40 & 43.20 & \best{57.92}\std{1.75} \\
        II          & 3 & 75.56 & 26.67 & 44.45 & 37.78 & 35.56 & 33.34 & \best{42.22}\std{2.22} & 65.33 & 57.33 & 61.33 & 54.67 & 32.00 & 54.13\std{3.95} \\
        III         & 4 & 71.11 & 22.22 & 37.78 & 35.56 & 35.56 & 31.11 & 38.89\std{5.88} & 62.67 & 62.67 & 60.00 & 57.33 & 44.00 & 57.33\std{5.33} \\
        IV          & 4 & 77.78 & 24.44 & 24.45 & 40.00 & 37.78 & 31.11 & 39.26\std{2.80} & 58.67 & 65.33 & 57.33 & 56.00 & 44.00 & 56.27\std{1.67} \\
        V           & 2 & 73.33 & 28.89 & 35.56 & 28.89 & 37.78 & 28.89 & 38.89\std{2.94} & 66.67 & 62.67 & 61.33 & 53.33 & 44.00 & 57.60\std{1.39} \\
        \bottomrule
        \end{tabular}
    }
    \vspace{2mm}
\end{table}
        
        \subsubsection{Role of scale space axes.}

            Since the scale-aware loss aligns the norm of the expected scale vector with experiential distance, the scale space may capture multiple aspects of situational novelty rather than a single one.
            We compare performance across 1D, 2D, and 3D scale spaces on EmbodiedBench using the L1 norm (Table~\ref{tab:scale_dim}).
            The 3D scale space yields the best average performance in both environments among the dimensionalities we evaluate, leading us to adopt it.
            
\begin{figure}[!tb]
    \centering
    \begin{subfigure}[t]{0.48\textwidth}
        \centering
        \begin{tikzpicture}
            \begin{axis}[
                scale only axis,
                width=0.75\textwidth, height=0.32\textwidth,
                xlabel={Scale},
                ylabel={Meta-routing score},
                xmin=0, xmax=1, ymin=0, ymax=4.9,
                legend style={at={(0.98,0.98)}, anchor=north east, font=\scriptsize, draw=none, fill=white, fill opacity=0.8, text opacity=1},
                grid=major, grid style={gray!20},
                tick label style={font=\scriptsize},
                label style={font=\small},
            ]
            \addplot[thick, color=red!80!black, smooth, domain=0:1, samples=80]
                {(1/0.4940998595614133)*exp(-((x-0.23584824800491333)^2)/(2*0.03885523974895477))};
            \addplot[thick, color=blue!80!black, smooth, domain=0:1, samples=80]
                {(1/0.2756097491222684)*exp(-((x-0.1891259253025055)^2)/(2*0.012089526280760765))};
            \addplot[thick, color=green!60!black, smooth, domain=0:1, samples=80]
                {(1/0.5713399897393815)*exp(-((x-0.39615869522094727)^2)/(2*0.051952850073575974))};
            \legend{Axis 1, Axis 2, Axis 3}
            \end{axis}
        \end{tikzpicture}
        \caption{Routine object manipulation in a familiar room. Axes concentrate toward low scales.}
        \label{fig:scale_axes_a}
    \end{subfigure}
    \hfill
    \begin{subfigure}[t]{0.48\textwidth}
        \centering
        \begin{tikzpicture}
            \begin{axis}[
                scale only axis,
                width=0.75\textwidth, height=0.32\textwidth,
                xlabel={Scale},
                yticklabels={},
                xmin=0, xmax=1, ymin=0, ymax=4.9,
                grid=major, grid style={gray!20},
                tick label style={font=\scriptsize},
                label style={font=\small},
            ]
            \addplot[thick, color=red!80!black, smooth, domain=0:1, samples=80]
                {(1/0.5436697675298257)*exp(-((x-0.1580955684185028)^2)/(2*0.04704251140356064))};
            \addplot[thick, color=blue!80!black, smooth, domain=0:1, samples=80]
                {(1/0.6214437996696749)*exp(-((x-0.20016580820083618)^2)/(2*0.06146442890167236))};
            \addplot[thick, color=green!60!black, smooth, domain=0:1, samples=80]
                {(1/0.470066477848177)*exp(-((x-0.8275551795959473)^2)/(2*0.03516727313399315))};
            \end{axis}
        \end{tikzpicture}
        \caption{Navigation to an unseen room with unfamiliar layout. Axis 3 moves to high scales while the others stay low.}
        \label{fig:scale_axes_b}
    \end{subfigure}

    \vspace{0.5mm}

    \begin{subfigure}[t]{0.48\textwidth}
        \centering
        \begin{tikzpicture}
            \begin{axis}[
                scale only axis,
                width=0.75\textwidth, height=0.32\textwidth,
                xlabel={Scale},
                ylabel={Meta-routing score},
                xmin=0, xmax=1, ymin=0, ymax=4.9,
                grid=major, grid style={gray!20},
                tick label style={font=\scriptsize},
                label style={font=\small},
            ]
            \addplot[thick, color=red!80!black, smooth, domain=0:1, samples=80]
                {(1/0.4653343974041719)*exp(-((x-0.9106276631355286)^2)/(2*0.03446279093623161))};
            \addplot[thick, color=blue!80!black, smooth, domain=0:1, samples=80]
                {(1/0.32165047701757515)*exp(-((x-0.7400404810905457)^2)/(2*0.016466015949845314))};
            \addplot[thick, color=green!60!black, smooth, domain=0:1, samples=80]
                {(1/0.23915383187460612)*exp(-((x-0.5735911726951599)^2)/(2*0.009102796204388142))};
            \end{axis}
        \end{tikzpicture}
        \caption{Multi-step task requiring sequential subgoal planning. Axes 1 and 2 move to high scales while Axis 3 stays lower.}
        \label{fig:scale_axes_c}
    \end{subfigure}
    \hfill
    \begin{subfigure}[t]{0.48\textwidth}
        \centering
        \begin{tikzpicture}
            \begin{axis}[
                scale only axis,
                width=0.75\textwidth, height=0.32\textwidth,
                xlabel={Scale},
                yticklabels={},
                xmin=0, xmax=1, ymin=0, ymax=4.9,
                grid=major, grid style={gray!20},
                tick label style={font=\scriptsize},
                label style={font=\small},
            ]
            \addplot[thick, color=red!80!black, smooth, domain=0:1, samples=80]
                {(1/0.29343487939529006)*exp(-((x-0.8776362538337708)^2)/(2*0.01370388176292181))};
            \addplot[thick, color=blue!80!black, smooth, domain=0:1, samples=80]
                {(1/0.10873009601096345)*exp(-((x-0.8860383629798889)^2)/(2*0.0018815669463947415))};
            \addplot[thick, color=green!60!black, smooth, domain=0:1, samples=80]
                {(1/0.21362100995227387)*exp(-((x-0.892932653427124)^2)/(2*0.007262866478413343))};
            \end{axis}
        \end{tikzpicture}
        \caption{First encounter with a completely novel environment. All axes shift to high scales.}
        \label{fig:scale_axes_d}
    \end{subfigure}
    \caption{\textbf{Per-axis meta-routing score distributions for four input types on EmbodiedBench.} Each curve shows $\wmr(\s)$ projected onto one axis of the 3D scale space.}
    \label{fig:scale_axes}
\end{figure}
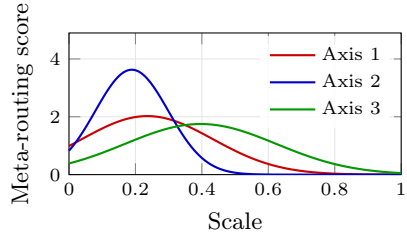
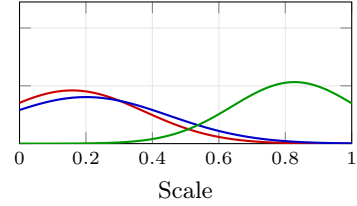
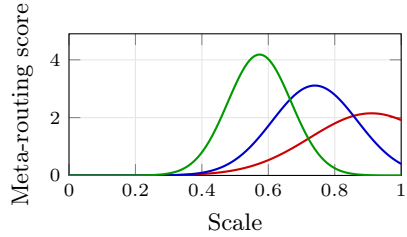
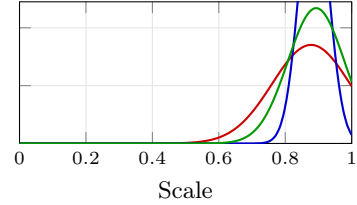

            To understand why additional dimensions help, we inspect how individual axes respond to different inputs (Figure~\ref{fig:scale_axes}).
            We find that the axes move toward low or high scales depending on the novelty of the current observation, sometimes converging across the three axes and sometimes separating.
            Notably, the scale-aware loss supervises only $\|\mu_{\mathcal{S}}\|$, yet the axes respond differently across situations rather than in unison, indicating that each axis comes to encode a distinct aspect of novelty without any axis-level supervision.
            This emergent specialization suggests that a higher-dimensional scale space offers the capacity for such structure to arise, accounting for the consistent gains we observe from higher dimensionality.
        
            \begin{table}[t]
    \centering
    \caption{\textbf{Effect of scale space dimensionality on EmbodiedBench.} We report success rate (\%) for each capability subset.}
    \label{tab:scale_dim}
    \footnotesize
    \setlength{\tabcolsep}{3pt}
    \centering
    \resizebox{\textwidth}{!}{
        \begin{tabular}{@{}lcccccc>{\columncolor{avgcol}}cccccc>{\columncolor{avgcol}}c@{}}
        \toprule
        \multirow{2}{*}[-0.5ex]{\textbf{Dim}} & \multicolumn{7}{c}{\textbf{EB-Habitat}} & \multicolumn{6}{c}{\textbf{EB-Navigation}} \\
        \cmidrule(lr){2-8} \cmidrule(lr){9-14}
        & Base & Cmn & Cpx & Vis & Spt & Lng & Avg ($\uparrow$) & Base & Cmn & Cpx & Vis & Lng & Avg ($\uparrow$) \\
        \midrule
        1D          & 86.67 & 13.33 & 20.00 & 40.00 & 40.00 & 33.33 & 38.89 & 68.00 & 48.00 & 56.00 & 60.00 & 36.00 & 53.60 \\
        2D          & 80.00 & 13.33 & 20.00 & 26.67 & 40.00 & 20.00 & 33.33 & 60.00 & 64.00 & 60.00 & 64.00 & 36.00 & 56.80 \\
        3D (default)& 73.33 & 26.67 & 37.33 & 42.67 & 36.00 & 26.67 & \best{40.44} & 63.20 & 63.20 & 57.60 & 62.40 & 43.20 & \best{57.92} \\
        \bottomrule
        \end{tabular}
    }
\end{table}
        
        \subsubsection{Experiential distance and world model group activation.}
        
            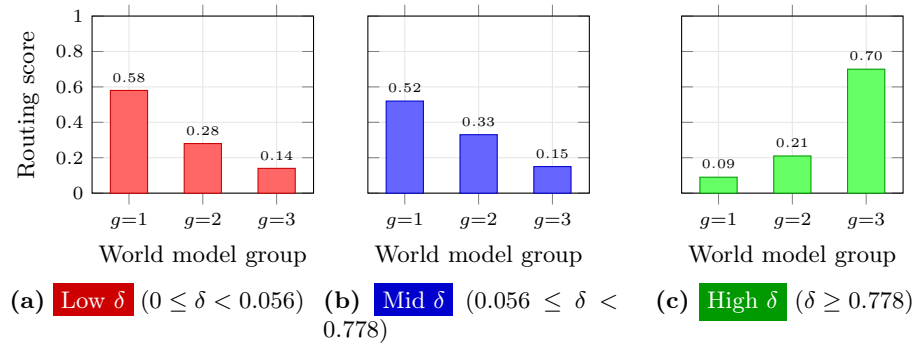
\begin{figure}[!tb]
    \centering
    \begin{subfigure}[t]{0.32\textwidth}
        \centering
        \begin{tikzpicture}
            \begin{axis}[
                scale only axis,
                width=0.75\textwidth, height=0.6\textwidth,
                ybar,
                bar width=14pt,
                xlabel={World model group},
                ylabel={Routing score},
                xmin=0.5, xmax=3.5,
                ymin=0, ymax=1.0,
                xtick={1,2,3},
                xticklabels={$g{=}1$, $g{=}2$, $g{=}3$},
                ytick={0, 0.2, 0.4, 0.6, 0.8, 1.0},
                grid=major, grid style={gray!20},
                tick label style={font=\scriptsize},
                label style={font=\small},
                nodes near coords,
                nodes near coords style={
                    /pgf/number format/fixed,
                    /pgf/number format/fixed zerofill,
                    /pgf/number format/precision=2,
                },
                every node near coord/.append style={font=\tiny},
            ]
            \addplot[fill=red!60!white, draw=red!80!black] coordinates {
                (1, 0.58) (2, 0.28) (3, 0.14)
            };
            \end{axis}
        \end{tikzpicture}
        \caption{\colorbox{red!80!black}{\textcolor{white}{Low $\exdist$}} ($0 \le \delta < 0.056$)}
        \label{fig:routing_low}
    \end{subfigure}
    \hfill
    \begin{subfigure}[t]{0.32\textwidth}
        \centering
        \begin{tikzpicture}
            \begin{axis}[
                scale only axis,
                width=0.75\textwidth, height=0.6\textwidth,
                ybar,
                bar width=14pt,
                xlabel={World model group},
                yticklabels={},
                xmin=0.5, xmax=3.5,
                ymin=0, ymax=1.0,
                xtick={1,2,3},
                xticklabels={$g{=}1$, $g{=}2$, $g{=}3$},
                ytick={0, 0.2, 0.4, 0.6, 0.8, 1.0},
                grid=major, grid style={gray!20},
                tick label style={font=\scriptsize},
                label style={font=\small},
                nodes near coords,
                nodes near coords style={
                    /pgf/number format/fixed,
                    /pgf/number format/fixed zerofill,
                    /pgf/number format/precision=2,
                },
                every node near coord/.append style={font=\tiny},
            ]
            \addplot[fill=blue!60!white, draw=blue!80!black] coordinates {
                (1, 0.52) (2, 0.33) (3, 0.15)
            };
            \end{axis}
        \end{tikzpicture}
        \caption{\colorbox{blue!80!black}{\textcolor{white}{Mid $\exdist$}} ($0.056 \le \delta < 0.778$)}
        \label{fig:routing_mid}
    \end{subfigure}
    \hfill
    \begin{subfigure}[t]{0.32\textwidth}
        \centering
        \begin{tikzpicture}
            \begin{axis}[
                scale only axis,
                width=0.75\textwidth, height=0.6\textwidth,
                ybar,
                bar width=14pt,
                xlabel={World model group},
                yticklabels={},
                xmin=0.5, xmax=3.5,
                ymin=0, ymax=1.0,
                xtick={1,2,3},
                xticklabels={$g{=}1$, $g{=}2$, $g{=}3$},
                ytick={0, 0.2, 0.4, 0.6, 0.8, 1.0},
                grid=major, grid style={gray!20},
                tick label style={font=\scriptsize},
                label style={font=\small},
                nodes near coords,
                nodes near coords style={
                    /pgf/number format/fixed,
                    /pgf/number format/fixed zerofill,
                    /pgf/number format/precision=2,
                },
                every node near coord/.append style={font=\tiny},
            ]
            \addplot[fill=green!60!white, draw=green!60!black] coordinates {
                (1, 0.09) (2, 0.21) (3, 0.70)
            };
            \end{axis}
        \end{tikzpicture}
        \caption{\colorbox{green!60!black}{\textcolor{white}{High $\exdist$}} ($\delta \ge 0.778$)}
        \label{fig:routing_high}
    \end{subfigure}
    \caption{\textbf{World model group activation at different experiential distances on EmbodiedBench.} Each bar shows the aggregate routing score for group $g$ ($g=1$: low-scale, $g=2$: mid-scale, $g=3$: high-scale). Low, Mid, and High denote equal-frequency tertiles of experiential distance $\delta$ (each $n \approx 790$).}
    \label{fig:routing}
\end{figure}
            We examine how experiential distance $\delta$ translates into world model group activation on EmbodiedBench (Figure~\ref{fig:routing}).
            We group samples into equal-frequency tertiles of $\delta$, so the bin boundaries follow the empirical distribution of $\delta$.
            Routing concentrates on the low-scale group across the \textcolor{red!80!black}{lower} and \textcolor{blue!80!black}{middle} tertiles ($\delta < 0.778$), with a slight shift of mass toward the mid-scale group in the middle tertile, and moves to the high-scale group once $\delta$ reaches the \textcolor{green!60!black}{top tertile} ($\delta \ge 0.778$).
            This shows that $\delta$ aligns world model activation with the appropriate scale, with familiar situations handled by low-scale models and novel ones routed to high-scale models.

\section{Conclusion} \label{sec:conclusion}

    We presented \oursol, a multi-scale mixture of world models framework that enables embodied agents to dynamically select and adapt world models across scales.
    By grounding routing in experiential distance, a measure of situational familiarity inspired by Construal Level Theory, our two-stage scale-based routing decomposes world model selection into scale determination and per-scale expert selection, making the scale of each decision explicit and amenable to targeted test-time updates.
    Our intra- and inter-scale knowledge adaptation mechanisms respect the distinct temporal characteristics at each scale, allowing low-level knowledge to update rapidly while preserving stable high-level abstractions, with gated cross-scale transfer maintaining coherence across the hierarchy.
    Experiments on EmbodiedBench and HAZARD demonstrate consistent improvements over state-of-the-art baselines in both multi-scale reasoning and dynamic adaptation scenarios, and analyses confirm that the framework generalizes across different world model architectures.

    \subsubsection{Limitations and future work.}

        Our framework inherits the capabilities and limitations of the underlying language model; reasoning quality is bounded by the VLM backbone, and improvements in base model capacity would directly benefit our approach.
        While our real-world experiments demonstrate initial transfer, broader validation across diverse robotic platforms and longer-horizon tasks would further establish the practical applicability of the approach.
        Promising future directions include automatically discovering the appropriate number and composition of world model groups from data, and extending real-world evaluation to a wider range of embodied platforms and task domains.


\section*{Acknowledgements}

    This work was supported by Institute of Information \& communications Technology Planning \& Evaluation (IITP) grant funded by the Korea government (MSIT),
    (RS-2022-II220043, Adaptive Personality for Intelligent Agents,
    RS-2022-II221045, Self-directed multi-modal Intelligence for solving unknown, open domain problems,
    RS-2025-02218768, Accelerated Insight Reasoning via Continual Learning,
    RS-2025-25442569, AI Star Fellowship Support Program (Sung-kyunkwan Univ.),
    RS-2026-25543726, Development of Leading Talent in Medical Domain-Specific Generative AI,
    RS-2026-25528384, Resource-Intensive AI Technologies Based on Sustainable GPU Integrated Platforms,
    RS-2019-II190421, Artificial Intelligence Graduate School Program (Sungkyunkwan University)),
    the National Research Foundation of Korea (NRF) grant funded by the Korea government (MSIT) (No. RS-2026-25474409),
    IITP-ITRC (Information Technology Research Center) grant funded by the Korea government (MSIT) (IITP-2025-RS-2024-00437633),
    and by Samsung Electronics Co., Ltd.

%
%
\bibliographystyle{styles/splncs04}
\bibliography{main}

\appendix
\onecolumn

\section{Benchmarks} \label{sec:benchmark}

    \subsection{EmbodiedBench} \label{subsec:embodied_bench}

        We conduct our experiments using EmbodiedBench~\cite{EmbodiedBench}, a comprehensive benchmark designed to evaluate vision-driven embodied agents powered by multi-modal large language models.
        It comprises 1{,}128 evaluation instances across four environments, covering tasks from high-level instruction following to low-level navigation and manipulation.
        In our experiments, we focus on the EB-Habitat and EB-Navigation environments.

        \begin{figure}[!tb]
    \centering
    \includegraphics[width=0.65\textwidth]{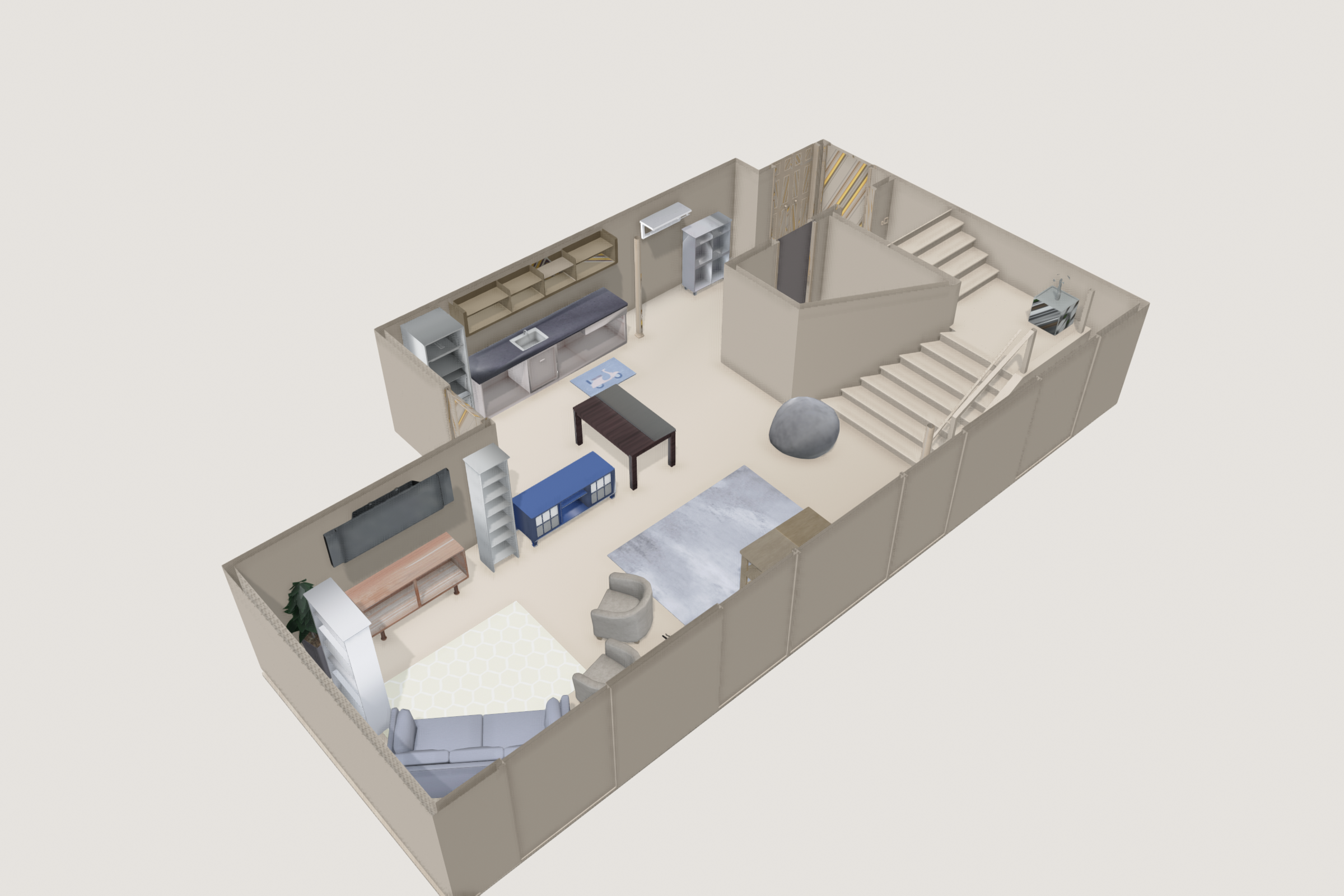}
    \caption{The top-view example of the Habitat environment.}
    \vspace{-0.08in}
    \label{fig:eb_habitat_top_view}
\end{figure}

        \subsubsection{Environment structure.}

            EmbodiedBench follows an episodic interaction protocol in which each instance provides a natural-language instruction and an initial environment configuration.
            At each step, the agent receives egocentric visual observations (along with textual feedback on action validity and optional auxiliary signals depending on the environment) and outputs an executable action plan.
            
        \subsubsection{Action space.}

            EmbodiedBench provides a hierarchical action space that bridges high-level semantic planning and low-level atomic control.
            In EB-Habitat, the action space consists of high-level skills (e.g., \texttt{navigate}, \texttt{pick}, \texttt{place}, \texttt{open}, \texttt{close}) parameterized by specific objects or receptacles.
            This formulation requires agents to focus on task decomposition and logical sequencing.
            In EB-Navigation, the action space is defined by discrete atomic commands directly executable by physical robots.
            It employs 8 discrete actions for translational movements (\texttt{forward}/\texttt{backward}/\texttt{left}/\texttt{right} by $\Delta x$), yaw rotations $(\Delta \theta)$, and camera tilt adjustments $(\Delta \varphi)$.

        \subsubsection{Task specifications.}

            To ensure rigorous assessment, we partition the dataset of the selected environments.
            In our experimental setup, the first 35 episodes of each subset are exclusively reserved as training data and a knowledge pool for retrieval-augmented in-context demonstrations, while the remaining unseen episodes (36 to 50 for EB-Habitat, and 36 to 60 for EB-Navigation) are used for evaluation.
            Task success is defined via PDDL-based goal conditions in EB-Habitat, and by a predefined distance threshold to the target in EB-Navigation.
            To evaluate diverse reasoning capabilities, tasks are structured into fine-grained subsets.
            EB-Habitat utilizes all six subsets (Base, Common Sense, Complex Instruction, Visual Appearance, Spatial Awareness, and Long Horizon), while EB-Navigation employs five, omitting spatial awareness.

    \subsection{HAZARD} \label{subsec:hazard}
    
        HAZARD~\cite{HAZARD} is a simulated embodied benchmark built on the ThreeDWorld platform, where agents must rescue valuable objects from dynamically evolving disaster scenarios.
        It features three disaster types---fire, flood, and wind---in which the environment spontaneously changes over time through spreading flames, rising water, or turbulent winds, demanding rapid perception, reasoning, and action from the agent.

        \begin{figure}[!tb]
    \centering
    \begin{subfigure}[t]{0.48\textwidth}
        \centering
        \centering
        \includegraphics[width=0.95\textwidth]{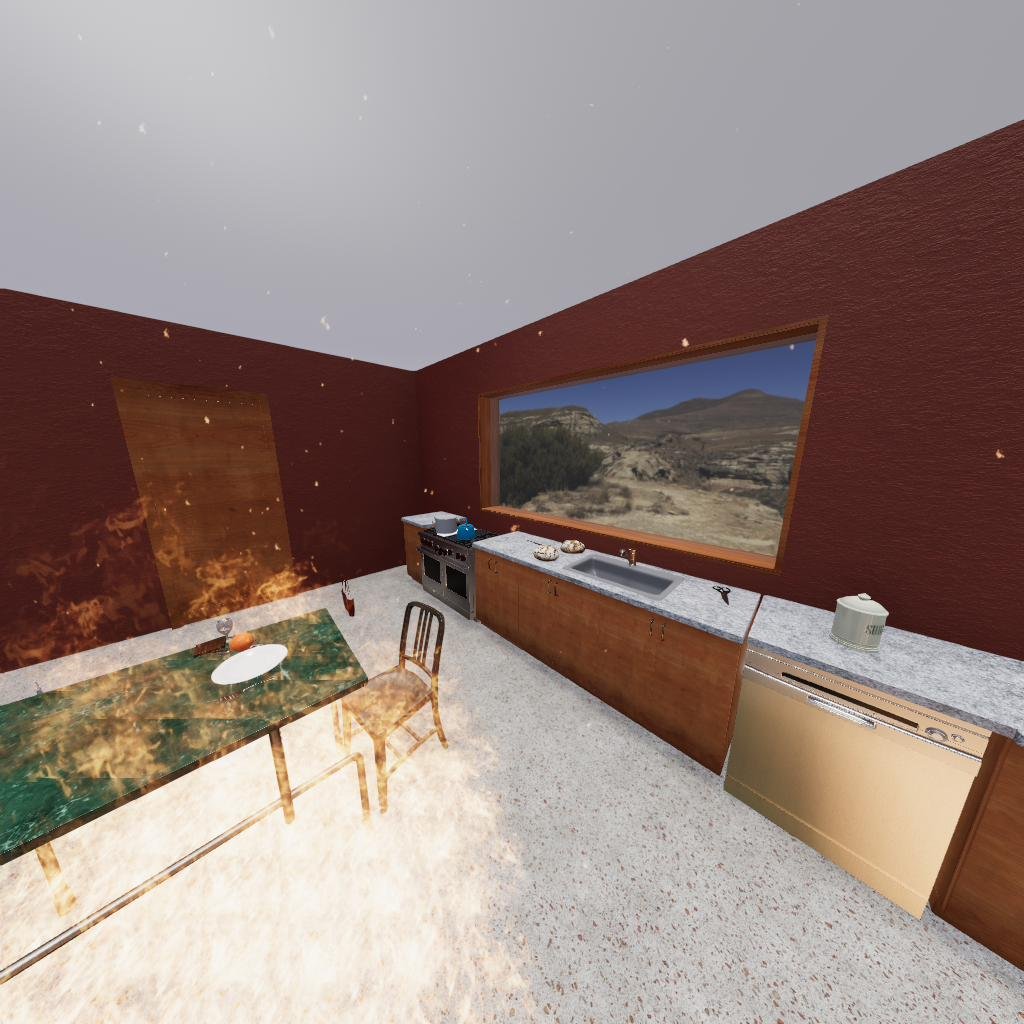}
        \caption{Fire}
        \label{fig:hazard_example_a}
    \end{subfigure}
    \hfill
    \begin{subfigure}[t]{0.48\textwidth}
        \centering
        \includegraphics[width=0.95\textwidth]{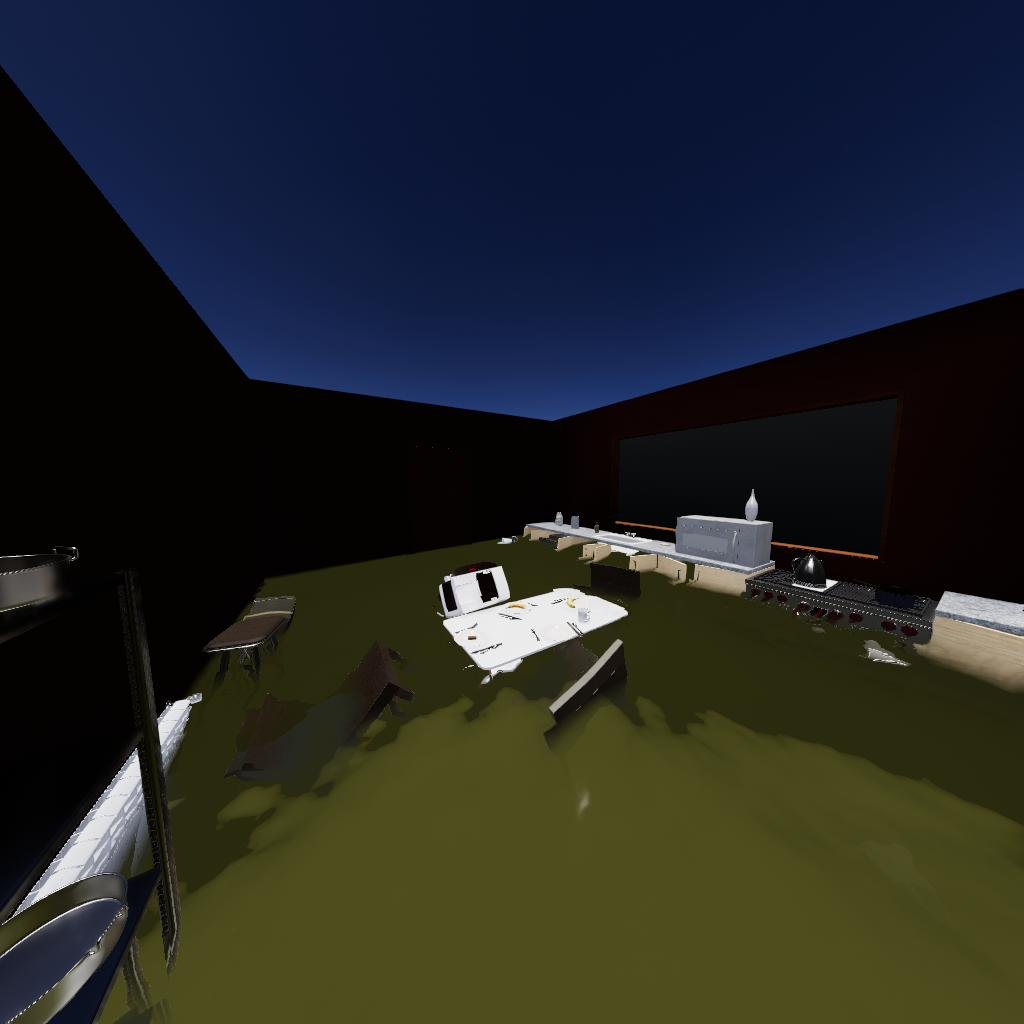}
        \caption{Flood}
        \label{fig:hazard_example_b}
    \end{subfigure}
    \caption{The example scene of the HAZARD environment.}
    \label{fig:hazard_examples}
\end{figure}
            
        \subsubsection{Environment structure.}

            At each step, the agent receives RGB-D observations, environment-specific signals such as temperature or water level, target object information (e.g., object names and associated values) along with an optional semantic segmentation mask depending on the evaluation setting.
            Each disaster scenario introduces distinct environmental dynamics.
            In fire, a temperature-based propagation system causes flames to spread and progressively destroy flammable objects.
            In flood, buoyancy and drag forces cause rising water to submerge and displace objects.
            In wind, aerodynamic forces scatter lightweight objects across the scene, complicating retrieval.
            We evaluate the performance of the baselines and \oursol in fire and flood scenario.
            
        \subsubsection{Action space.}

            HAZARD provides four high-level compressed actions derived from low-level primitives: \texttt{Walk To}, \texttt{Pick Up}, \texttt{Drop}, and \texttt{Explore}.
            The \texttt{Walk To} action leverages A*-based path planning to compress multiple navigation primitives into a single command, significantly reducing the frequency of LLM queries.
            As a result, the LLM-based decision maker only needs to select which object to walk to, while \texttt{Pick Up} and \texttt{Drop} handle object manipulation, and \texttt{Explore} enables the agent to survey its surroundings.
            
        \subsubsection{Task specifications.}

            The objective for the agent is to rescue a predetermined set of target objects by transporting them to designated safe locations, such as a bag or a shopping cart.
            Performance is evaluated using three metrics.
            The rescued value rate (Value) captures the ratio of rescued object value to total target value.
            The averaged rescue step (Step) measures the efficiency of the rescue process.
            The averaged damaged rate (Damage) tracks the proportion of rescued objects that sustained damage from environmental hazards.

\section{Implementation Details} \label{sec:implementation}
        
    \subsection{\oursol} \label{implementation_oursol}

        Tables~\ref{tab:hyperparams_mmow_train} and~\ref{tab:hyperparams_mmow_infer} summarize the training and inference hyperparameters for \oursol.
        The knowledge states described in Section~\ref{sec:adaptation} are implemented as a test-time training (TTT) memory module that maintains per-group knowledge states $\knowledge^{(\group)}$ as updatable parameter vectors.
        At every inference step, this module realizes both intra-scale adaptation (scale-dependent forgetting) and inter-scale adaptation (gated bidirectional transfer across neighboring groups).
        After each action execution, the agent computes a knowledge increment from the prediction error and distributes it across groups according to the meta-router weights, while the gated transfer matrices propagate relevant information between adjacent scales to maintain hierarchical coherence.

        \begin{table}[!tb]
    \centering
    \caption{Training hyperparameters for \oursol.}
    \label{tab:hyperparams_mmow_train}
    \begin{tabular}{l c}
        \toprule
        \textbf{Hyperparameter} & \textbf{Value} \\
        \midrule
        \multicolumn{2}{l}{\textit{Architecture}} \\
        \quad Base Model                & Qwen3-VL-4B-Instruct \\
        \quad World Model Types          & PINN, Cognitive Map, Relational \\
        \quad Num World Models           & 15 (3 types $\times$ 5) \\
        \quad Routing Top-$k$           & 4 \\
        \quad LoRA Rank / $\alpha$ / Dropout & 16 / 32 / 0.05 \\
        \quad Embedding Dim             & 384 \\
        \quad Scale Dim                 & 3 \\
        \quad MC Samples                & 16 \\
        \quad Target Encoder            & CLIP-ViT-Large-Patch14 \\
        \midrule
        \multicolumn{2}{l}{\textit{Optimization}} \\
        \quad Learning Rate             & $1 \times 10^{-4}$ \\
        \quad Optimizer                 & AdamW \\
        \quad Weight Decay              & 0.01 \\
        \quad Batch Size                & 1 \\
        \quad Num Epochs                & 1 \\
        \quad LR Schedule               & Cosine \\
        \quad Warmup Steps              & 200 \\
        \quad Max Grad Norm             & 1.0 \\
        \midrule
        \multicolumn{2}{l}{\textit{Loss Weights}} \\
        \quad $\lambda_{\text{action}}$ / $\lambda_{\text{aux}}$ / $\lambda_{\text{transition}}$ & 0.1 / 0.1 / 1.0 \\
        \quad Scale-Aware Loss ($\lambda_S$) & 0.1 \\
        \quad \quad Variance reg.\ ($\lambda_1$) / $Z$ reg.\ ($\lambda_2$) & 0.01 / 0.01 \\
        \quad Diversity Loss ($\lambda$) & 0.01 \\
        \quad \quad Dormant Boost / Threshold / EMA & 2.0 / 0.1 / 0.99 \\
        \midrule
        \multicolumn{2}{l}{\textit{Curriculum Phases}} \\
        \quad Phase A (router only)     & steps 0--500 \\
        \quad Phase B (+world models)    & steps 500--1{,}500 \\
        \quad Phase C (+diversity)      & steps 1{,}500+ \\
        \bottomrule
    \end{tabular}
\end{table}
        \begin{table}[!tb]
    \centering
    \caption{Inference hyperparameters for \oursol.}
    \label{tab:hyperparams_mmow_infer}
    \begin{tabular}{l c}
        \toprule
        \textbf{Hyperparameter} & \textbf{Value} \\
        \midrule
        Max New Tokens              & 50 \\
        Temperature                 & 1.0 \\
        \midrule
        \multicolumn{2}{l}{\textit{Knowledge Adaptation}} \\
        \quad $\alpha_{\max}$       & 0.3 \\
        \quad $\gamma$              & 1.0 \\
        \bottomrule
    \end{tabular}
\end{table}

        \subsubsection{World Model Architectures.} \label{subsubsec:world_model_architectures}

            Our framework employs seven types of world model architectures, each designed to capture a distinct aspect of environment dynamics.
            All world models share the same LoRA adapter configuration for the base VLM backbone but differ in their auxiliary prediction heads and loss functions.
            Table~\ref{tab:world_model_aux_loss} summarizes the auxiliary loss weight for each architecture.

            \paragraph{Sensory.}
            The Sensory world model encodes low-level affordance signals by predicting three binary properties from the VLM's hidden state $\mathbf{h}_{\text{state}}$: whether the agent is near an interactable object, whether the target is pickable, and whether a placement location is available.
            Each property is predicted by an independent two-layer MLP ($d_{\text{model}} \to d_{\text{model}}/4 \to 1$, GELU, dropout).
            The auxiliary loss sums three binary cross-entropy terms: $\mathcal{L}_{\text{Sensory}} = \text{BCE}(\hat{y}_{\text{near}}, y_{\text{near}}) + \text{BCE}(\hat{y}_{\text{pick}}, y_{\text{pick}}) + \text{BCE}(\hat{y}_{\text{place}}, y_{\text{place}})$.

            \paragraph{Schema.}
            The Schema world model captures action precondition rules by predicting whether an action is valid and, if not, the reason for failure.
            It consists of two heads: a validity head ($d_{\text{model}} \to d_{\text{model}}/4 \to 1$, GELU, dropout) producing a binary logit, and a failure-reason head ($d_{\text{model}} \to d_{\text{model}}/4 \to 5$) classifying among five failure categories.
            The auxiliary loss is $\mathcal{L}_{\text{Schema}} = \text{BCE}(\hat{v}, v) + \text{CE}(\hat{f}, f)$, where the failure-reason term is computed only over invalid-action samples.

            \paragraph{Concept Bottleneck.}
            The Concept Bottleneck world model predicts interpretable state concepts from $\mathbf{h}_{\text{state}}$.
            Three parallel two-layer MLPs predict the currently held object (34 classes), the open/closed state of receptacles (5-dim binary), and the current location (16 classes), respectively.
            The auxiliary loss is $\mathcal{L}_{\text{Concept}} = \text{CE}(\hat{h}, h) + \text{BCE}(\hat{o}, o) + \text{CE}(\hat{l}, l)$.

            \paragraph{PINN.}
            The Physics-Informed Neural Network (PINN) world model learns an explicit state transition function $g_\theta(\hat{s}_t, a_t) \to \hat{s}_{t+1}$.
            The environment state is decomposed into the same three components as the Concept Bottleneck (held object, open state, location), but instead of predicting them from $\mathbf{h}_{\text{state}}$, PINN concatenates one-hot encodings of the current state and action to form a 60-dimensional input.
            This input is passed through a two-layer MLP ($60 \to d_h/4 \to d_h/4$, GELU, dropout) followed by three parallel linear heads.
            The auxiliary loss has the same form: $\mathcal{L}_{\text{PINN}} = \text{CE}(\hat{h}, h) + \text{BCE}(\hat{o}, o) + \text{CE}(\hat{l}, l)$.

            \paragraph{Cognitive Map.}
            The Cognitive Map world model focuses on spatial structure by predicting location-level outcomes from $\mathbf{h}_{\text{state}}$.
            It consists of two parallel two-layer MLPs ($d_{\text{model}} \to d_{\text{model}}/4 \to 16$, GELU, dropout), one predicting the next location and the other the target location.
            The auxiliary loss is $\mathcal{L}_{\text{CM}} = \text{CE}(\hat{l}_{\text{next}}, l_{\text{next}}) + \text{CE}(\hat{l}_{\text{target}}, l_{\text{target}})$.

            \paragraph{RBF Network.}
            The Radial Basis Function (RBF) world model learns a contrastive embedding space that distinguishes successful from unsuccessful action executions.
            A two-layer MLP ($d_{\text{model}} \to d_{\text{model}}/4 \to 256$, GELU, dropout) projects $\mathbf{h}_{\text{state}}$ into an $\ell_2$-normalized 256-dimensional embedding.
            The auxiliary loss is an InfoNCE objective with temperature $\tau{=}0.07$, where positive pairs share a successful action execution and negative pairs share the same action type but differ in success: $\mathcal{L}_{\text{RBF}} = -\log \frac{\exp(\text{sim}(\mathbf{z}, \mathbf{z}^+)/\tau)}{\sum_j \exp(\text{sim}(\mathbf{z}, \mathbf{z}_j)/\tau)}$.

            \paragraph{Relational Network.}
            The Relational world model reasons over pairwise object interactions using a relation network architecture.
            Unlike the other world models, it operates on frozen CLIP embeddings rather than the VLM's hidden state.
            For each pair of objects $(i, j)$ in an episode (up to 49 objects), a pairwise relation function $g_\theta([\mathbf{e}_i; \mathbf{e}_j; \mathbf{a}_t; \mathbf{u}_t])$ produces a relation vector $\mathbf{r}_{ij} \in \mathbb{R}^{512}$, where $\mathbf{e}$ are CLIP object embeddings and $\mathbf{a}_t$, $\mathbf{u}_t$ are CLIP-encoded action and instruction strings.
            An attention pooling layer, with query derived from $[\mathbf{z}_t; \mathbf{a}_t; \mathbf{u}_t]$, aggregates over all valid pairs to produce a context vector $\mathbf{h}_t$.
            A prediction head then maps $[\mathbf{z}_t; \mathbf{a}_t; \mathbf{h}_t]$ to a predicted next-observation embedding $\hat{\mathbf{z}}_{t+1}$.
            The loss combines cosine and $\ell_2$ objectives: $\mathcal{L}_{\text{RN}} = \alpha (1 - \cos(\hat{\mathbf{z}}, \mathbf{z}_{t+1})) + (1 - \alpha) \|\hat{\mathbf{z}} - \mathbf{z}_{t+1}\|_2^2$ with $\alpha = 0.5$.
            Since the Relational loss directly serves as the transition prediction objective, no additional auxiliary loss weighting is applied.

            \begin{table}[t]
    \centering
    \caption{Auxiliary loss weights ($\lambda$) for each world model architecture. All scheduled losses use linear warmup over the first 10\% of training steps. The Relational Network loss is used directly as the transition prediction objective without additional weighting.}
    \label{tab:world_model_aux_loss}
    \begin{tabular}{l c c}
        \toprule
        \textbf{World Model} & $\boldsymbol{\lambda}$ & \textbf{Scheduling} \\
        \midrule
        Sensory             & 0.10 & Linear warmup \\
        Schema              & 0.20 & Linear warmup \\
        Concept Bottleneck  & 0.20 & Linear warmup \\
        PINN                & 0.05 & Linear warmup \\
        Cognitive Map       & 0.20 & Linear warmup \\
        RBF Network         & 0.10 & Linear warmup \\
        Relational Network  & --   & -- \\
        \bottomrule
    \end{tabular}
\end{table}

    \subsection{Baselines} \label{subsec:implementation_baselines}

        \subsubsection{LLM-Planner.}
            
            LLM-Planner retrieves relevant demonstration snippets based on the current observation and instruction, and concatenates them with the planner prompt as few-shot examples for the model to propose the next action.
            We include this baseline to compare retrieval-augmented in-context learning directly with \oursol.
            In our implementation, we replace the text-only LLM with a vision-language model to enable direct visual grounding from egocentric observations.
            Table~\ref{tab:hyperparams_llmplanner} summarizes the inference hyperparameters.

            \begin{table}[!tb]
    \centering
    \caption{Hyperparameters for LLM-Planner (inference only).}
    \label{tab:hyperparams_llmplanner}
    \begin{tabular}{l c}
        \toprule
        \textbf{Hyperparameter} & \textbf{Value} \\
        \midrule
        Base Model          & Qwen3-VL-4B-Instruct \\
        Max New Tokens      & 1{,}024 \\
        Temperature         & 1.0 \\
        Top-$p$             & 0.95 \\
        RAG Top-$k$         & 5 \\
        Retrieval           & BERT emb.\ + obs.\ rerank \\
        Max Plan Length     & No limit \\
        Precision           & bfloat16 \\
        Special Mechanism   & Failed action filtering \\
        \bottomrule
    \end{tabular}
\end{table}

        \subsubsection{SayCanPay.}

            SayCanPay combines LLM-based task planning with a learned cost model to select grounded and cost-aware actions.
            It consists of three components, where the Say model proposes candidate actions, the Can model provides environmental affordances, and the Pay model estimates execution costs via reinforcement learning.
            We include this baseline to compare learning-based action scoring with \oursol.
            In our implementation, we fine-tune Qwen3-VL-4B-Instruct with LoRA adapters for both the Say and Pay models jointly.
            Tables~\ref{tab:hyperparams_saycanpay_train} and~\ref{tab:hyperparams_saycanpay_infer} summarize the training and inference hyperparameters, respectively.

            \begin{table}[!tb]
    \centering
    \caption{Training hyperparameters for SayCanPay.}
    \label{tab:hyperparams_saycanpay_train}
    \begin{tabular}{l c}
        \toprule
        \textbf{Hyperparameter} & \textbf{Value} \\
        \midrule
        Base Model              & Qwen3-VL-4B-Instruct \\
        Model Type              & Both (Say + Pay) \\
        LoRA Rank / $\alpha$    & 16 / 32 \\
        Learning Rate           & $2 \times 10^{-4}$ \\
        Batch Size              & 4 \\
        Num Epochs              & 1 \\
        Max Sequence Length      & 1{,}024 \\
        Discount Factor (Pay)   & 0.9 \\
        Quantization            & Off \\
        \bottomrule
    \end{tabular}
\end{table}
            \begin{table}[!tb]
    \centering
    \caption{Inference hyperparameters for SayCanPay.}
    \label{tab:hyperparams_saycanpay_infer}
    \begin{tabular}{l c}
        \toprule
        \textbf{Hyperparameter} & \textbf{Value} \\
        \midrule
        Temperature             & 1.0 (nucleus) \\
        Top-$p$                 & 0.6 \\
        Pay Top-$k$             & 30 \\
        Scoring                 & $\log P_{\text{say}} + \log P_{\text{pay}}$ (token-level) \\
        Max Plan Length          & Auto-regressive \\
        Precision               & bfloat16 \\
        Special Mechanism       & Action blocking \\
        \bottomrule
    \end{tabular}
\end{table}

        \subsubsection{FLARE.}

            FLARE enhances standard retrieval-based planning by incorporating a dynamic replanning mechanism.
            By monitoring the current scene, it adjusts action sequences on the fly.
            Specifically, if an action fails due to a mismatched object name, the agent identifies and substitutes alternative targets using semantic similarity.
            We include this baseline to compare \oursol against a representative retrieval-based method with adaptive replanning capabilities.
            Table~\ref{tab:hyperparams_flare} summarizes the inference hyperparameters.

            \begin{table}[!tb]
    \centering
    \caption{Hyperparameters for FLARE (inference only).}
    \label{tab:hyperparams_flare}
    \begin{tabular}{l c}
        \toprule
        \textbf{Hyperparameter} & \textbf{Value} \\
        \midrule
        Base Model          & Qwen3-VL-4B-Instruct \\
        Max New Tokens      & 1{,}024 \\
        Temperature         & 1.0 \\
        Top-$p$             & 0.95 \\
        RAG Top-$k$         & 5 \\
        Retrieval           & Multi-Modal (BERT + CLIP, $w_l{=}0.5$, $w_e{=}0.5$) \\
        Max Plan Length      & 10 \\
        Precision           & bfloat16 \\
        Special Mechanism   & Env.\ Adaptive Replanning (EAR) \\
        \bottomrule
    \end{tabular}
\end{table}

        \subsubsection{Conventional MoE.}

            To isolate the contribution of our proposed expert routing strategy, we implement a conventional Mixture-of-Experts baseline that employs the identical backbone architecture as \oursol.
            This baseline attaches seven LoRA adapters as experts to Qwen3-VL-4B-Instruct, with a learned router that assigns input tokens to experts based on standard top-$k$ gating.
            By comparing against this baseline, we evaluate whether the structured expert decomposition in \oursol provided advantages over a generic MoE formulation.
            Tables~\ref{tab:hyperparams_moe_train} and~\ref{tab:hyperparams_moe_infer} summarize the training and inference hyperparameters, respectively.

            \begin{table}[!tb]
    \centering
    \caption{Training hyperparameters for Conventional MoE.}
    \label{tab:hyperparams_moe_train}
    \begin{tabular}{l c}
        \toprule
        \textbf{Hyperparameter} & \textbf{Value} \\
        \midrule
        \multicolumn{2}{l}{\textit{Architecture}} \\
        \quad Base Model                & Qwen3-VL-4B-Instruct \\
        \quad Num Experts                & 7 \\
        \quad Top-$k$                    & 3 \\
        \quad LoRA Rank / $\alpha$       & 16 / 32 \\
        \quad Expert Dropout             & 0.2 \\
        \quad Router $d_{\text{model}}$  & 512 \\
        \quad Router Num Heads / Aspects & 8 / 8 \\
        \quad Router FFN Dim             & 1{,}024 \\
        \quad Router Dropout             & 0.1 \\
        \midrule
        \multicolumn{2}{l}{\textit{Optimization}} \\
        \quad Learning Rate (LoRA)       & $2 \times 10^{-4}$ \\
        \quad Learning Rate (Router)     & $1 \times 10^{-3}$ \\
        \quad Batch Size                 & 1 \\
        \quad Gradient Accum.\ Steps     & 32 (eff.\ batch = 32) \\
        \quad Num Epochs                 & 1 \\
        \quad Warmup Ratio               & 0.05 \\
        \quad Max Grad Norm              & 1.0 \\
        \quad Max Seq Length              & 4{,}096 \\
        \quad Image Size                 & $256 \times 256$ \\
        \quad Quantization               & 4-bit \\
        \midrule
        \multicolumn{2}{l}{\textit{Loss Weights}} \\
        \quad LM Loss                    & 1.0 \\
        \quad Router LM Loss             & 1.0 \\
        \quad Balance Loss               & $0.1 \to 0.05$ (annealing) \\
        \bottomrule
    \end{tabular}
\end{table}
            \begin{table}[!tb]
    \centering
    \caption{Inference hyperparameters for Conventional MoE.}
    \label{tab:hyperparams_moe_infer}
    \begin{tabular}{l c}
        \toprule
        \textbf{Hyperparameter} & \textbf{Value} \\
        \midrule
        Max New Tokens      & 1{,}024 \\
        Temperature         & 1.0 \\
        Top-$p$             & 0.95 \\
        Max Plan Length      & No limit \\
        Image Resize        & $256 \times 256$ \\
        Precision           & bfloat16 \\
        Scoring             & Router $\to$ LoRA expert $\to$ VLM gen. \\
        \bottomrule
    \end{tabular}
\end{table}

\section{Additional Experiments} \label{sec:additional_experiments}

    \subsubsection{Experiential distance analysis.}
        We replace the Mahalanobis distance with cosine similarity and KL divergence.
        For KL, we fit the embedding distribution of experience $\mathbf{E}_{<t}$ as $\mathcal{N}(\boldsymbol{\mu}_E, \boldsymbol{\Sigma}_E)$ and treat the current observation as $\mathcal{N}(\boldsymbol\phi(o_t), \mathbf{I})$, using $\mathrm{KL}\!\left(\mathcal{N}(\boldsymbol\phi(o_t), \mathbf{I}) \,\|\, \mathcal{N}(\boldsymbol{\mu}_E, \boldsymbol{\Sigma}_E)\right)$.
        \begin{table}[!tb]
    \centering
    \caption{Experiential distance analysis.}
    \label{tab:add_exp_dis_anal}
    \footnotesize
    \setlength{\tabcolsep}{3pt}
    \centering
    \begin{tabular}{lcc}
        \toprule
        Distance metric & SR ($\uparrow$) & PS ($\downarrow$) \\
        \midrule
        Cosine similarity & 35.11\std{3.00} & 10.30\std{1.12} \\
        KL divergence & 38.22\std{5.70} & 7.44\std{0.60} \\
        \textbf{Mahalanobis (ours)} & \best{40.44}\std{1.27} & 7.25\std{0.50} \\
        \bottomrule
    \end{tabular}
\end{table}
        
        As shown in Table~\ref{tab:add_exp_dis_anal}, Mahalanobis is \textbf{+2.22\,\%p} ahead of the next best.
        Cosine is unfavorable for $\mathcal{L}_\mathcal{S}^{(1)}$ alignment in three ways: it yields a non-convex objective in $\boldsymbol\phi(o_t)$, ignores embedding magnitude, and cannot exploit the experience covariance $\boldsymbol\Sigma_E$ (all directions are treated equally).
        Under this direction, KL contains the (squared) Mahalanobis distance as its mean-difference term, which explains why it tracks Mahalanobis more closely than cosine.
        
    \subsubsection{Compute cost.}
        
        \begin{table}[!tb]
    \centering
    \caption{Compute cost comparison.}
    \label{tab:compute_cost}
    \footnotesize
    \setlength{\tabcolsep}{3pt}
    \centering
    \begin{tabular}{lcccc}
        \toprule
        & LLM-Planner & SayCanPay & Conv.\ MoE & \textbf{MuSix} \\
        \midrule
        Lat.\ (s) & 2.00 & 1.86 & 2.52 & \textbf{3.52} \\
        Mem.\ (GB) & 8.30 & 8.43 & 8.43 & \textbf{9.60} \\
        \bottomrule
    \end{tabular}
\end{table}
    
        We profile \oursol against baselines on EB-Habitat (RTX Pro 6000 Blackwell, batch=1).
        The two-stage routing and 16-sample MC integration add \textbf{39.7}\% latency over conventional MoE (Table~\ref{tab:compute_cost}), in exchange for the reported accuracy gains.

\end{document}